%% file: paper.tex
\documentclass{article}

\PassOptionsToPackage{numbers}{natbib}
\usepackage[preprint]{neurips_2026}

\usepackage[utf8]{inputenc}
\usepackage[T1]{fontenc}
\usepackage{hyperref}
\usepackage{url}
\usepackage{booktabs}
\usepackage{amsfonts}
\usepackage{nicefrac}
\usepackage{microtype}
\usepackage{xcolor}
\usepackage{colortbl}
\definecolor{posblue}{RGB}{210,228,242}
\definecolor{negred}{RGB}{246,217,217}
\usepackage{graphicx}
\usepackage{subcaption}
\usepackage{caption}
\usepackage{amsmath}
\usepackage{amssymb}
\usepackage{mathtools}
\usepackage{amsthm}
\usepackage{tabularx}
\usepackage{multirow}
\usepackage{float}
\usepackage{placeins}  
\usepackage{enumitem}
\setlist{noitemsep}

\usepackage[capitalize,noabbrev]{cleveref}

\theoremstyle{plain}

\theoremstyle{definition}

\theoremstyle{remark}

\usepackage[disable,textsize=tiny]{todonotes}

\makeatletter
\newcommand{\appendixtableofcontents}{%
  \section*{Appendix Contents}%
  \@starttoc{atoc}%
  \clearpage
}
\newcommand{\addappendixtocline}[2]{%
  \addtocontents{atoc}{\protect\contentsline{#1}{#2}{\thepage}{}}%
}
\newcommand{\enableappendixtoc}{%
  \let\originalappendixsection\section
  \let\originalappendixsubsection\subsection
  \renewcommand{\section}[1]{%
    \originalappendixsection{##1}%
    \addappendixtocline{section}{\protect\numberline{\thesection}##1}%
  }%
  \renewcommand{\subsection}[1]{%
    \originalappendixsubsection{##1}%
    \addappendixtocline{subsection}{\protect\numberline{\thesubsection}##1}%
  }%
}
\newcommand{\disableappendixtoc}{%
  \let\section\originalappendixsection
  \let\subsection\originalappendixsubsection
}
\makeatother

\title{SyntaxBench: A Statistical Diagnostic Framework for Character-Level Reasoning in Large Language Models}

\newcommand{\protectedemail}[2]{\raisebox{-#1}{\includegraphics{#2}}}
\author{%
  Mohsen Larni\thanks{These authors contributed equally.} \\
  Department of Computer Science \\
  University of Nevada, Las Vegas \\
  \protectedemail{0.83313pt}{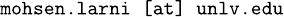} \\
  \And
  Sobhan Ebrahimi Azar\footnotemark[1] \\
  Department of Computer Science \\
  University of Nevada, Las Vegas \\
  \protectedemail{0.83313pt}{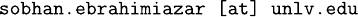} \\
  \AND
  Pouyan Nahed \\
  Department of Computer Science \\
  University of Nevada, Las Vegas \\
  \protectedemail{2.22168pt}{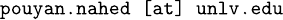} \\
  \And
  Kazem Taghva \\
  Department of Computer Science \\
  University of Nevada, Las Vegas \\
  \protectedemail{2.22168pt}{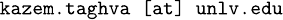} \\
}

\begin{document}

\maketitle

\begin{abstract}
\input{files/0_abstract}
\end{abstract}

\section{Introduction}\label{sec:intro}
\input{files/1_introduction}

\section{Related Work}\label{sec:related}
\input{files/2_related_works}

\section{Methodology}\label{sec:methodology}
\input{files/3_methodology}

\section{Experiments and Results}\label{sec:experiments}
\input{files/4_experiments_and_results}

\section{Discussion}\label{sec:discussion}
\input{files/5_discussion}

\section{Limitations}\label{sec:limitations}
\input{files/6_limitations}

\section{Conclusion}\label{sec:conclusion}
\input{files/8_conclusion}

\bibliographystyle{plain}
\bibliography{example_paper}

\newpage
\appendix
\appendixtableofcontents
\enableappendixtoc
\input{files/appendix/methodology}
\input{files/appendix/tasks}
\disableappendixtoc

\end{document}

%% file: files/0_abstract.tex
Large language models are increasingly used in settings where small syntactic errors matter, yet character-level reasoning is still evaluated mostly through isolated probes and aggregate accuracy. We introduce \textbf{SyntaxBench}, a diagnostic benchmark and statistical evaluation framework for character-level reasoning. SyntaxBench contains five core deterministic tasks, character counting, letter containment, palindrome detection, edit distance, and longest-string selection, plus \texttt{index\_to\_span}, a harder substring-extraction stress test. The five core tasks are instantiated on paired English and character-length-matched random-string inputs. \texttt{index\_to\_span} documents are drawn from the same 200--500 word band and are not character-length matched. All six tasks use zero-, one-, and four-shot prompts.

We evaluate eight open-weight models spanning 2B--32B parameters across 11 available reasoning-mode configurations. The framework reports exact-match and relaxed accuracy together with Cohen's $\kappa$, paired McNemar tests with odds-ratio effect sizes, bootstrap confidence intervals, Kendall's $\tau$ over shot count, class-conditional metrics, tokenization analysis, and multiple-comparison-corrected condition tests. Three findings stand out. First, tokenization strongly shapes accuracy: random strings are more character-visible than English strings (1.892 vs.\ 3.169 chars/token), and \texttt{count\_chars} accuracy falls sharply as English words occupy more tokens. Second, reasoning mode is not uniformly helpful: Gemma4-31B is essentially indistinguishable across modes on the near-saturated tasks, while Qwen3.6-27B is worse with thinking enabled on \texttt{palindrome\_test} (0.952 non-thinking vs.\ 0.886 thinking at four-shot). Third, \texttt{index\_to\_span} remains largely unsolved; the best four-shot exact-match accuracy is 6.75\%.

SyntaxBench shows that character-level evaluation needs controlled input conditions, paired statistical tests, and analyses of tokenization and reasoning mode rather than aggregate accuracy alone. The code will be released soon.

\textbf{Keywords:} Large Language Models, Character-Level Reasoning, Tokenization, Chain-of-Thought, Diagnostic Benchmark, Statistical Evaluation

%% file: files/1_introduction.tex
Large language models (LLMs) are now used as everyday tools for writing, programming, tutoring, and analysis. In these settings, users often expect exact answers to small syntactic questions: How many characters are in a string? Does a letter appear? What is the edit distance between two strings? These operations are simple for deterministic programs, but they remain brittle for LLMs. The familiar ``strawberry'' counting failure~\cite{shin2024large,bostrom2024characters} is a visible instance of a broader problem: fluent text generation does not imply reliable character-level reasoning.

Character-level tasks are useful because they have deterministic ground truth. They also expose failure modes that semantic benchmarks can hide. A model can sound confident while being off by one character, returning the wrong binary label, or extracting the wrong substring from a long context. Standard benchmarks~\cite{wang2018glue,wang2019superglue,srivastava2022beyond,liang2023holistic} measure broad language understanding, but they rarely isolate these low-level operations. When character-level probes are used, results are often summarized with aggregate accuracy alone, without paired tests, effect sizes, class-conditional behavior, or controls for tokenization and input distribution.

We introduce \textbf{SyntaxBench}, a diagnostic benchmark and evaluation framework for this setting. The benchmark has six deterministic tasks. Five are core character-level tasks: \texttt{count\_chars}, \texttt{letter\_container}, \texttt{palindrome\_test}, \texttt{edit\_distance}, and \texttt{longest\_string\_finder}. The sixth, \texttt{index\_to\_span}, is a harder stress test that asks models to recover a substring from explicit start and end indices in a long text. The five core tasks are instantiated on paired English and length-matched random-string inputs, allowing us to separate linguistic familiarity from character-level processing. The stress-test documents are drawn from the same 200--500 word band and are not character-length matched. We evaluate zero-, one-, and four-shot prompting, and compare thinking and non-thinking modes where both are available.

SyntaxBench is organized around four research questions. \textbf{RQ1 (mechanism):} is the English-vs-random accuracy difference driven by tokenizer granularity (characters per token) rather than linguistic familiarity? \textbf{RQ2 (reasoning mode):} treated as a deployment switch, when does enabling thinking help, hurt, or leave character-level accuracy unchanged? \textbf{RQ3 (short-to-long transfer):} does competence on short-input character operations transfer to character-offset extraction over long documents? \textbf{RQ4 (methodology):} what statistical machinery is needed to make character-level benchmark claims falsifiable rather than anecdotal?

The framework is designed for benchmark interpretation, not just score reporting. In addition to exact-match and relaxed accuracy, it reports bootstrap confidence intervals, Cohen's $\kappa$, paired McNemar tests with odds-ratio effect sizes, Kendall's $\tau$ over shot count, class-conditional binary metrics, tokenization-stratified accuracy, integer and string error summaries, and multiple-comparison-corrected English-vs-random tests. These diagnostics make each headline claim falsifiable: they distinguish model differences from sample noise, input-condition effects from tokenization artifacts, and accuracy gains from validity or class-bias effects.

We evaluate eight open-weight models spanning 2B--32B parameters: the Qwen3.5 family at 2B, 9B, and 27B; Qwen3.6-27B; Gemma4-31B; GPT-OSS-20B; Nemotron-Cascade-2; and DeepSeek-R1-32B. The resulting 11 available model-mode configurations show that character-level reasoning is not a single capability. Several models saturate \texttt{letter\_container} and perform well on \texttt{palindrome\_test}, while \texttt{edit\_distance} remains strongly model-dependent and \texttt{index\_to\_span} is near the floor for every model. At four-shot in thinking mode, GPT-OSS-20B has the best composite score (macro EMA 0.818), but even it reaches only 6.75\% exact match on \texttt{index\_to\_span}.

Three empirical patterns motivate the benchmark. \textbf{First}, tokenization mediates English-vs-random behavior. Across tokenizer-covered models, random strings are more character-visible than English strings: 0.542 versus 0.344 tokens per character, or 1.892 versus 3.169 characters per token (Cohen's $d = 2.165$ on tokens per character). On \texttt{count\_chars}, accuracy declines as English examples occupy more tokens, consistent with errors inside multi-character subwords. \textbf{Second}, reasoning mode is not uniformly helpful. DeepSeek-R1-32B benefits from thinking on several tasks, but Qwen3.6-27B is worse with thinking enabled on \texttt{palindrome\_test} (0.952 non-thinking vs.\ 0.886 thinking at four-shot). \textbf{Third}, scale helps but does not solve the benchmark. Within Qwen3.5, four-shot \texttt{count\_chars} improves from 0.698 to 0.880 to 0.958 across 2B, 9B, and 27B, yet the complete 9B and 27B \texttt{index\_to\_span} cells remain at $0.5\%$, and the 2B 4-shot cell has no random-condition responses (0/113 English).

\textbf{Contributions.} We make three contributions. (1) \emph{SyntaxBench}: a six-task diagnostic benchmark for character-level reasoning with paired English and random inputs. The five core tasks are character-length matched. The stress test matches word count, not character length. (2) \emph{A reproducible evaluation framework}: a pipeline that reports accuracy, validity, agreement, effect sizes, monotonicity, tokenization effects, class-conditional behavior, and multiple-comparison-corrected statistical tests. (3) \emph{An empirical study of open-weight LLMs}: an analysis showing tokenization-mediated input effects, non-uniform reasoning-mode behavior, within-family scaling, and a persistent failure mode on positional substring extraction. The code will be released soon.

%% file: files/2_related_works.tex
\paragraph{General-purpose LLM benchmarks.}
GLUE~\cite{wang2018glue}, SuperGLUE~\cite{wang2019superglue}, BIG-Bench~\cite{srivastava2022beyond}, MMLU~\cite{hendrycks2021measuring}, and HELM~\cite{liang2023holistic} established standardized evaluation suites for natural language understanding and broad reasoning. These benchmarks emphasize semantic capabilities and probe character-level operations only incidentally. SyntaxBench addresses this gap with deterministic ground truth, balanced positive/negative cases per task, and an evaluation framework that pairs accuracy with statistical machinery (paired McNemar, Cohen's $\kappa$, Kendall's $\tau$, tokenizer stratification).

\paragraph{Character-level reasoning, tokenization, and the ``strawberry'' problem.}
Subword tokenization decouples a model's input representation from individual characters, with measurable consequences for orthographic tasks. Tokenization-free architectures such as ByT5~\cite{xue2022byt5} and CANINE~\cite{clark2022canine} demonstrate improved robustness on character-level tasks but are not the dominant deployment paradigm. Recent work~\cite{shin2024large,bostrom2024characters} has shown that even frontier models miscount characters in common words. Our results in \S\ref{sec:tokenization} mechanistically link this failure mode to per-token chars-per-token packing rather than to lexical familiarity, showing that for tasks requiring per-character enumeration the gap is in the direction of \emph{tokenizer-induced opacity on linguistic input}, not unfamiliarity with random characters.

\paragraph{In-context learning and shot-mode effects.}
Brown et al.~\cite{brown2020language} introduced few-shot in-context learning; chain-of-thought prompting~\cite{wei2022chain} extended this to multi-step reasoning. Subsequent work has documented that few-shot performance is non-monotonic on certain tasks~\cite{lu2022fantastically,chen2023relation} and that demonstration selection~\cite{liu2022what} substantially affects outcomes. SyntaxBench operationalizes monotonicity testing via Kendall's $\tau$ over shot index per (model, mode, task, condition) and demonstrates that the same model on the same task can have sign-flipped trajectories between thinking and non-thinking modes (\S\ref{sec:diagnostic}).

\paragraph{Reasoning mode and overthinking.}
The reasoning-mode toggle has become standard in mid-sized open-weight LLMs (Qwen3.6, Gemma4, GPT-OSS, DeepSeek-R1). The overthinking literature~\cite{nye2021show} has documented that chain-of-thought traces can degrade performance on tasks where the underlying decision is direct. Our results in \S\ref{sec:thinking} extend this to character-level operations and provide a paired-McNemar test on the same model, showing the regime is not uniform: thinking can help, hurt, or be a no-op depending on (model, task, shot).

\paragraph{Algorithmic and symbolic reasoning evaluations.}
Scratchpads~\cite{nye2021show}, MATH~\cite{hendrycks2021math}, and GSM8K~\cite{cobbe2021gsm8k} target multi-step numeric reasoning rather than primitive string operations. SyntaxBench targets the lower stratum: deterministic operations whose ground truth is computable in a single pass, with partial-correctness metrics that decouple ``approximately right'' from ``exactly right''. The \texttt{index\_to\_span} stress test (\S\ref{sec:floor}) provides a long-horizon counter task that the core suite intentionally avoids.

\paragraph{Statistical methodology in LLM benchmarking.}
A recurring methodological concern is the reliance on point-estimate accuracy without effect sizes, confidence intervals, or correction for multiple comparisons~\cite{ulmer2022experimental,Miller2024Adding}. Established tools, McNemar's test~\cite{mcnemar1947note,dietterich1998approximate} for paired classifier comparisons, Cohen's $\kappa$~\cite{cohen1960coefficient} for class-imbalance-aware agreement, Benjamini-Hochberg FDR~\cite{benjamini1995controlling} for power-aware multiple correction, and lm-evaluation-harness~\cite{eval-harness} for standardized protocols, are rarely combined in a single benchmark release. SyntaxBench couples its dataset with a reproducible statistical framework that applies these tools together to surface findings invisible in accuracy alone.

\paragraph{Linguistic priors and generalization.}
Several studies show LMs exploit distributional properties of natural language at the expense of robust symbolic reasoning~\cite{mccoy2019right,niven2019probing}. The SyntaxBench length-matched random-string corpus for the five core tasks tests the ``LLMs degrade on non-natural inputs'' hypothesis directly: if degradation were universal, we would expect significant English-vs-random gaps across all tasks. Our chi-square analysis (\S\ref{sec:tokenization}) refines this picture: of the 25 Bonferroni-significant cells, 23 go the other direction (random easier) and 2 do not, and the mechanism on the majority is tokenizer granularity, not familiarity.

%% file: files/3_methodology.tex
\subsection{Benchmark Overview}\label{sec:benchmark}

SyntaxBench evaluates whether LLMs can perform deterministic syntactic operations under strict output constraints. The benchmark consists of \textbf{six tasks} (Table~\ref{tab:task_summary}): a five-task \emph{core} suite with interpretable outputs and well-defined partial-credit metrics, plus one \emph{stress test} (\texttt{index\_to\_span}) that locates a target substring at specified character offsets within a 200--500 word document. Each task has deterministic ground truth and is instantiated under three \emph{shot} modes (zero-, one-, four-shot) and, where supported, two \emph{thinking} modes (thinking, non-thinking), giving up to six (shot, thinking-mode) configurations per (model, task). Per-condition sample counts and prompt templates are in Appendix~\ref{app:tasks}.

\begin{table}[t]
\centering
\caption{SyntaxBench tasks. The \emph{Inputs} column reports per-condition sample counts (English / random).}
\label{tab:task_summary}
\small
\setlength{\tabcolsep}{4pt}
\begin{tabular}{@{}llcc@{}}
\toprule
\textbf{Task} & \textbf{Description} & \textbf{Output} & \textbf{Inputs (En / Rand)} \\
\midrule
\multicolumn{4}{l}{\emph{Intra-word core}} \\
\quad \texttt{count\_chars}        & length of a string $s$                           & integer & 200 / 200 \\
\quad \texttt{letter\_container}   & does $s$ contain character $c$?                  & yes/no  & 400 / 400 \\
\quad \texttt{palindrome\_test}    & does $s$ read the same forward and backward?     & yes/no  & 400 / 400 \\
\midrule
\multicolumn{4}{l}{\emph{Inter-word core}} \\
\quad \texttt{edit\_distance}      & Levenshtein distance between $(s_1, s_2)$        & integer & 200 / 200 \\
\quad \texttt{longest\_string\_finder} & longest of $\geq 3$ strings                  & string  & 200 / 200 \\
\midrule
\multicolumn{4}{l}{\emph{Stress test}} \\
\quad \texttt{index\_to\_span}     & substring $d[i:j]$ in long text                  & string  & 200 / 200 \\
\bottomrule
\end{tabular}
\end{table}

\subsection{Data Sources and Random-String Construction}

The English vocabulary is drawn from the NLTK corpus (236{,}736 entries). \texttt{palindrome\_test} additionally uses a curated 229-entry palindrome word list from Wiktionary, and \texttt{index\_to\_span} uses Wikipedia article excerpts. For the five core tasks, the \textbf{length-matched random-string corpus} matches the empirical character-length distribution of the NLTK vocabulary: for each English word of length $\ell$, a random string of the same length is drawn uniformly from $[a\text{-}z]$. This controls for the trivial confound that random strings might differ in length from English words, contaminating the English-vs-random comparison via length effects. \texttt{index\_to\_span} is the exception at the character level. Both conditions are targeted to 200--500 words (observed means 321 English and 336 random). English documents are Wikipedia excerpts. Random documents are assembled by sampling that many strings from the random vocabulary, not by replacing each English word in an excerpt. Those strings follow the NLTK word-list lengths rather than the short function words of Wikipedia prose, so mean document length is 3{,}593 characters on the random condition and 2{,}128 on the English condition.

\subsection{Models and Thinking Modes}\label{sec:models}

We evaluate \textbf{eight open-weight LLMs spanning a 16$\times$ parameter range across eleven (model, thinking-mode) configurations}: \textbf{Qwen3.5-2B / 9B / 27B}~\cite{qwen35} (non-thinking only); \textbf{Qwen3.6-27B}~\cite{qwen36} (both modes, vLLM at FP8); \textbf{Gemma4-31B-it}~\cite{gemma4} (both modes, vLLM at BF16, TP=2); \textbf{GPT-OSS-20B}~\cite{gptoss} (thinking, Ollama at MXFP4, the only stable backend at evaluation time); \textbf{Nemotron-Cascade-2-31B}~\cite{nemotron} (thinking, \texttt{llama.cpp} with split GGUF); and \textbf{DeepSeek-R1-32B}~\cite{deepseek} (thinking-by-design; its ``non-thinking'' configuration disables reasoning-trace \emph{accounting} rather than reasoning itself, and is reported separately). Two models (GPT-OSS-20B and Nemotron-Cascade-2) are therefore evaluated in thinking mode only. The Qwen3.5 family at three scales gives a within-family scaling curve at fixed architecture; Qwen3.6-27B and Gemma4-31B in both modes each provide a \emph{within-model} reasoning-mode ablation at fixed scale (the two models are distinct architectures). Per-(model, mode) backend tuning of \texttt{seq\_len} and \texttt{num\_seq} is recorded in the run manifest and reproduced in Appendix~\ref{app:experimental_setup}.

\subsection{Evaluation Protocol}

Each (model, thinking-mode) is evaluated under \textbf{0-, 1-, and 4-shot} prompting with deterministic decoding (temperature 0). The system prompt is task-specific; demonstrations (when present) are alternating user--assistant message pairs preceding the test query. Each test instance is an independent inference request reusing the system prompt and demonstrations but no prior model outputs. Models return a JSON object with a single \texttt{result} field; the response parser handles whitespace and JSON-formatting variability, including a Markdown-bold (\texttt{**\{...\}**}) fallback for one model whose chat template wraps responses in bold markers. Reasoning traces (when emitted in a separate field) are captured for diagnostics but excluded from scoring.

\subsection{Scoring and Metrics}\label{sec:scoring}

\paragraph{Primary metrics} per (model, task, shot, thinking-mode, condition) cell: Exact-Match Accuracy (EMA, parse errors counted as incorrect); Relaxed Accuracy (RA, $\pm 1$ tolerance, integer tasks); Parse Error Rate (PER); Cohen's $\kappa$~\cite{cohen1960coefficient} (unweighted for binary tasks, linearly-weighted for integer tasks).

\paragraph{Statistical inference.} (i) bootstrap 95\% CIs on per-task EMA via 10{,}000 resamples of samples; the composite macro-EMA interval instead resamples the six task-level EMAs (Appendix~\ref{app:stats}); (ii) paired McNemar~\cite{mcnemar1947note} for between-model and thinking-vs-non-thinking comparisons, with continuity correction, discordant counts $(b_{01}, b_{10})$, and odds ratio as effect size; (iii) chi-square tests of independence~\cite{pearson1900x} for English-vs-random per (model, task, shot, mode), with Cram\'er's $V$; (iv) Kendall's $\tau$~\cite{kendall1938new} between shot index and EMA. Multiple-comparison correction uses Bonferroni for headline tables and Benjamini--Hochberg FDR~\cite{benjamini1995controlling} for the supplementary $q$-values, both at $\alpha = 0.05$.

\paragraph{Tokenizer-stratified analysis.} For each model with a public HuggingFace tokenizer, we tokenize the raw input content and bin samples by token count, reporting mean chars-per-token by condition, Cohen's $d$ between English and random tokens-per-character distributions, and per-bucket accuracy. Method details and per-model tokenizer assignments are in Appendix~\ref{app:tokenization}.

%% file: files/4_experiments_and_results.tex
\subsection{Setup and Main Results}\label{sec:setup}

We evaluate the eight models from \S\ref{sec:models} on all six SyntaxBench tasks under 0-, 1-, and 4-shot prompting. The 11 (model, thinking-mode) configurations are those in \S\ref{sec:models}. Inference uses deterministic decoding (temperature 0); per-(model, mode) backend tuning is in Appendix~\ref{app:experimental_setup}.

\begin{figure}[t]
    \centering
    \includegraphics[width=0.85\linewidth]{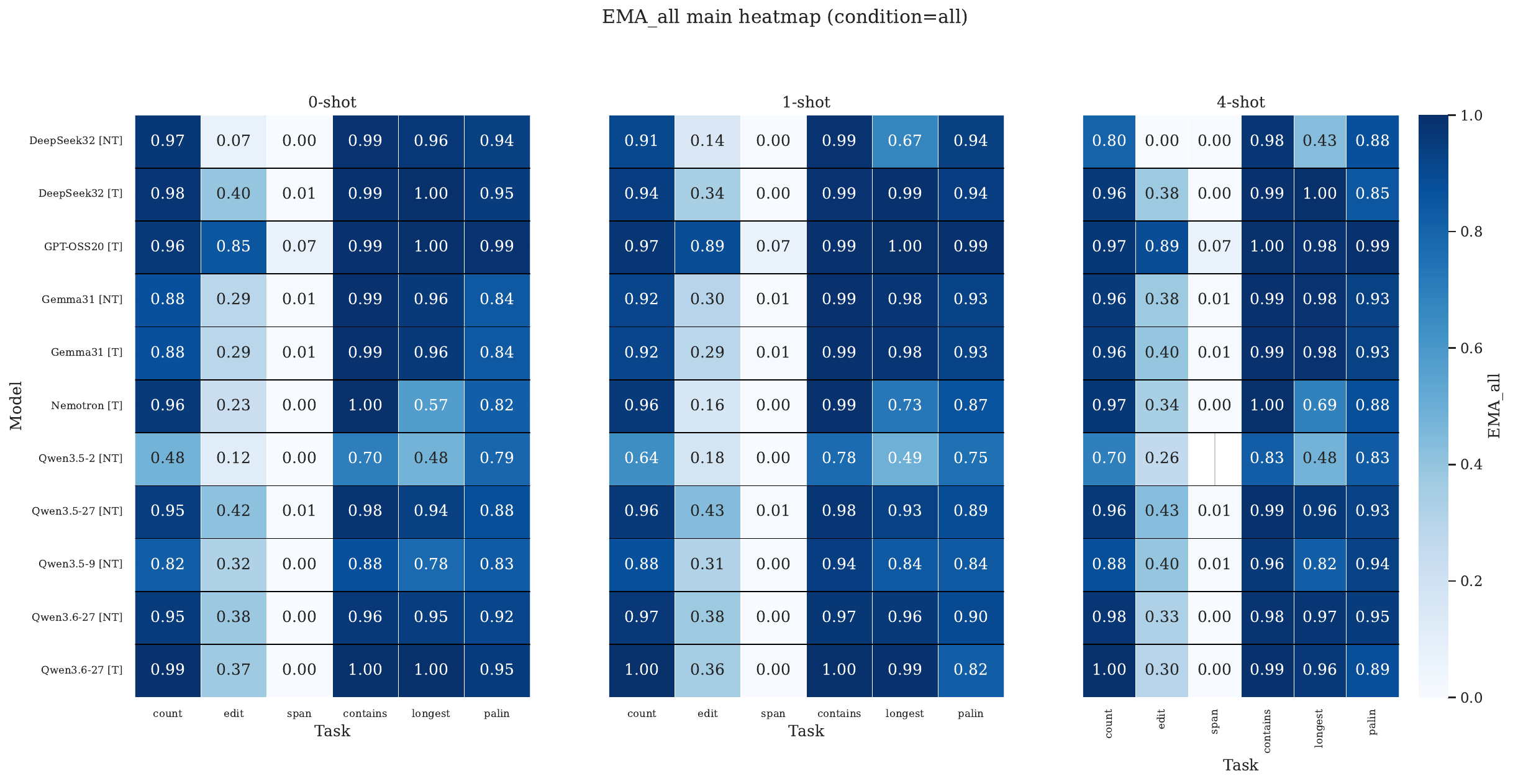}
    \caption{\textbf{Main results.} Per-(model, mode) EMA on combined English+random inputs at 0-, 1-, and 4-shot. At 4-shot, 25 of 32 cells on \texttt{count\_chars}, \texttt{letter\_container}, \texttt{longest\_string\_finder}, and \texttt{palindrome\_test} for 27B+ configurations reach 90\% EMA. Nemotron-Cascade-2 is at 69.0\% on \texttt{longest\_string\_finder}, and DeepSeek-R1-32B non-thinking is at 43.25\% on that task. \texttt{edit\_distance} discriminates models, and \texttt{index\_to\_span} remains near the floor. Empty cells are configurations that were not run, or a combined score that is not estimable because one input condition is missing.}
    \label{fig:main_heatmap}
\end{figure}

Figure~\ref{fig:main_heatmap} reports per-shot EMA on combined English+random inputs across all 11 configurations. Three patterns dominate the surface view: \emph{(i)} on \texttt{count\_chars}, \texttt{letter\_container}, \texttt{longest\_string\_finder}, and \texttt{palindrome\_test}, 25 of 32 4-shot cells for 27B+ configurations are at least 90\% EMA; the exceptions include Nemotron-Cascade-2 at 69.0\% combined on \texttt{longest\_string\_finder} and DeepSeek-R1-32B non-thinking at 43.25\% combined on that task; \emph{(ii)} \texttt{edit\_distance} discriminates strongly, with GPT-OSS-20B at $\mathbf{89.0\%}$ versus $0.0$--$43.0\%$ for the rest (the top of that range is Qwen3.5-27B non-thinking; DeepSeek-R1-32B non-thinking is $0.0\%$); \emph{(iii)} \texttt{index\_to\_span} is unsolved, with max 4-shot EMA of $\mathbf{6.75\%}$ (GPT-OSS-20B). Table~\ref{tab:composite_leaderboard} summarizes the 4-shot composite leaderboard; the per-task 4-shot table with 95\% bootstrap CIs and Bonferroni-corrected significance is in Appendix~\ref{app:full_results}, Table~\ref{tab:main_results}.

\begin{table}[t]
\centering
\caption{Composite leaderboard at 4-shot (thinking, condition=all). Macro EMA averages over the six tasks. The interval resamples those six task-level EMAs; it is not the sample bootstrap used for per-task EMA. VRR = valid-response rate; PER = parse-error rate. Qwen3.5 family entries (non-thinking-only) appear in Appendix~\ref{app:tables_main}, Table~\ref{tab:composite_leaderboard_nonthinking}.}
\label{tab:composite_leaderboard}
\small
\begin{tabular}{rlrcc}
\toprule
\textbf{Rank} & \textbf{Model} & \textbf{Macro EMA [95\% CI]} & \textbf{VRR} & \textbf{PER} \\
\midrule
1 & GPT-OSS-20B          & $0.818$ [0.512, 0.989] & 0.964 & 0.036 \\
2 & Gemma4-31B           & $0.712$ [0.388, 0.971] & 1.000 & 0.000 \\
3 & DeepSeek-R1-32B      & $0.697$ [0.385, 0.966] & 0.973 & 0.027 \\
4 & Qwen3.6-27B          & $0.690$ [0.358, 0.972] & 1.000 & 0.000 \\
5 & Nemotron-Cascade-2-31B & $0.648$ [0.339, 0.919] & 0.992 & 0.008 \\
\bottomrule
\end{tabular}
\end{table}

The wide CIs are diagnostic: macro EMA hides the floor effect on \texttt{index\_to\_span}, the discrimination gap on \texttt{edit\_distance}, and the saturated near-ceiling behavior on the other four tasks. The next four subsections break the surface view apart along the three abstract findings (tokenization, reasoning mode, and the \texttt{index\_to\_span} floor) and a fourth subsection that demonstrates the diagnostic value of the framework itself.

\subsection{Headline 1: Tokenization Mediates the English-vs-Random Gap}\label{sec:tokenization}

The natural reading of LLM degradation on non-linguistic input is lexical familiarity: a model has not seen \texttt{xqfb} but has seen \texttt{cat}. We test this directly via chi-square independence tests on each (model, task, shot, thinking-mode) cell. After Bonferroni correction across the 197 comparisons in which both conditions were evaluated (11 configurations $\times$ 6 tasks $\times$ 3 shots, excluding the incomplete Qwen3.5-2B \texttt{index\_to\_span} 4-shot cell), \textbf{25 (12.7\%) show a significant English-vs-random gap}. Random inputs are easier in 23 of those cells. English is easier in the other two: DeepSeek-R1-32B on \texttt{longest\_string\_finder} at 4-shot non-thinking, and Gemma4-31B on \texttt{edit\_distance} at 0-shot thinking. The mechanism on the random-easier majority is tokenization, not familiarity.

\begin{figure}[t]
    \centering
    \includegraphics[width=0.95\linewidth]{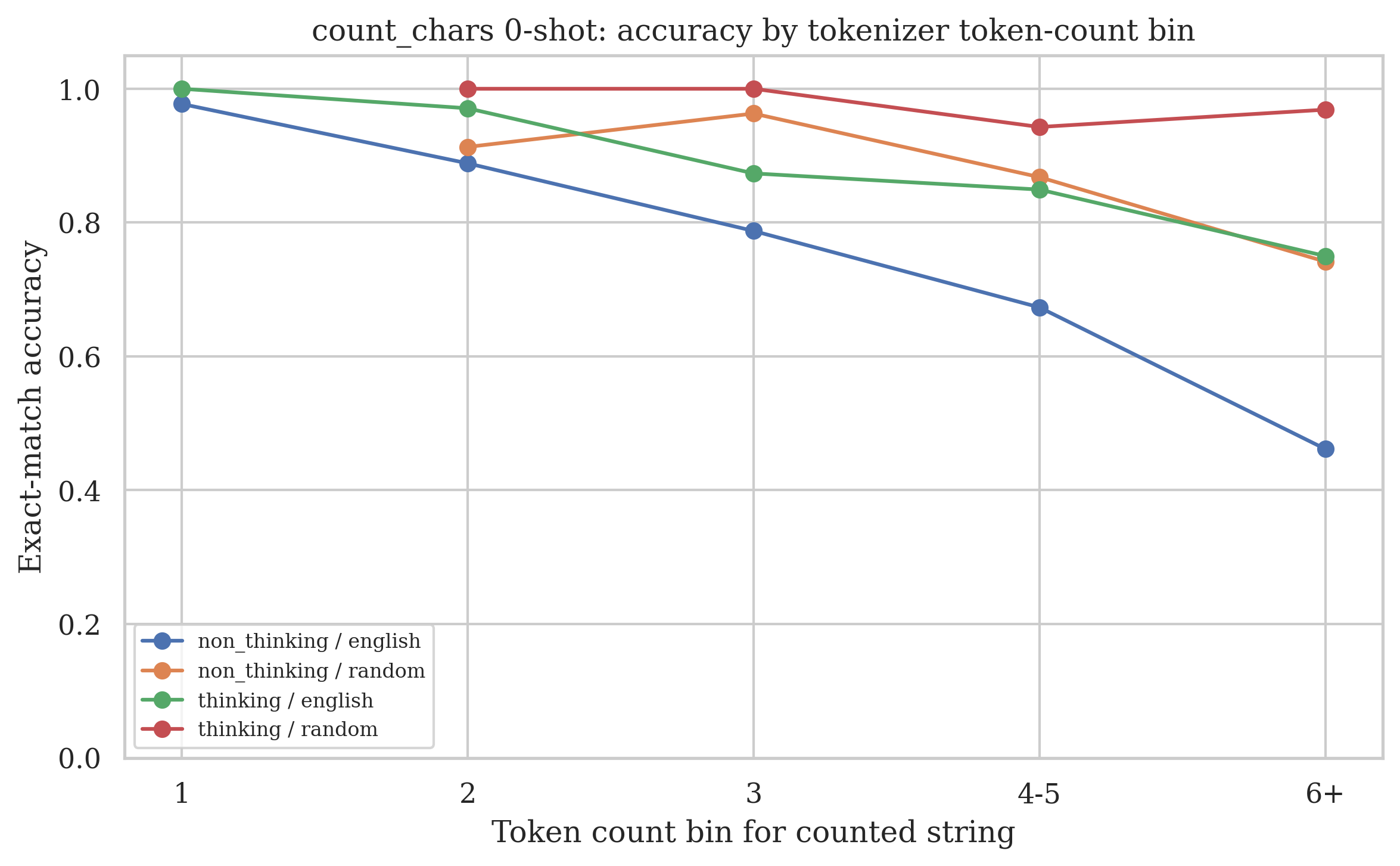}
    \caption{\textbf{0-shot \texttt{count\_chars} accuracy by tokenizer-token count.} The figure shows this task only. English accuracy falls from $97.7\%$ to $46.2\%$ across the 1-token to 6+ bins in non-thinking mode, and from $100\%$ to $75\%$ in thinking mode. Random inputs remain higher than English in every shared bin.}
    \label{fig:token_acc}
\end{figure}

Applying each model's HuggingFace tokenizer to raw input strings on the five core tasks, and excluding the single-character letter query, English packs $3.169$ chars/token while random packs $1.892$ chars/token, a $1.67\times$ ratio. On tokens per character the same samples give $0.344$ versus $0.542$, with Cohen's $d = 2.165$. Stratifying 0-shot accuracy by input token count (Figure~\ref{fig:token_acc}), \texttt{count\_chars} English accuracy falls from $97.7\%$ to $46.2\%$ across the 1-token to 6+ bins in non-thinking mode, and from $100\%$ to $75\%$ in thinking mode. Random inputs remain higher than English in every shared bin. Per-example failure analysis confirms the mechanism: nearly every off-by-one English failure on \texttt{count\_chars} localizes to multi-character subwords. Gemma4-31B miscounts \texttt{undeclared} (true 10) as 11 with tokenization \texttt{unde|cl|ared}; \texttt{osphresiology} (true 13) as 14 with \texttt{osph|res|iology}; \texttt{verisimilarly} (true 13) as 14 with \texttt{ver|isim|ilarly}. The non-thinking 1-token English bin is correct on 43 of 44 items (97.7\%). \texttt{getter} and \texttt{terminal} are examples from that bin.

The implication for benchmark design is sharp: the standard ``LLMs fail on non-linguistic input'' framing conflates lexical familiarity and tokenizer granularity. Tokenizer-stratified evaluation disentangles them, and on character-level tasks the gap is in the direction of \emph{tokenizer-induced opacity on linguistic input}, not unfamiliarity with random characters. The tokenizer-stratified method is in Appendix~\ref{app:tokenization}. Per-bucket accuracies and per-tokenizer statistics are in the released tokenization CSVs, not tabulated in the appendix.

\subsection{Headline 2: Reasoning Mode Helps Unevenly and Sometimes Hurts}\label{sec:thinking}

For Qwen3.6-27B and Gemma4-31B we evaluate both thinking and non-thinking configurations of the same model, and DeepSeek-R1-32B is thinking-by-design. The naive prediction is that thinking dominates non-thinking. Our data refute this for character-level tasks.

\begin{table}[t]
\centering
\caption{Qwen3.6-27B on \texttt{palindrome\_test}, EMA and Cohen's $\kappa$ on combined inputs. Non-thinking outperforms thinking at 1-shot and 4-shot, with the gap concentrated in the positive class (at 4-shot, TPR drops from $0.9075$ to $0.7725$ and TNR goes from $0.9975$ to $1.0000$).}
\label{tab:thinking_palindrome}
\small
\begin{tabular}{lcccccc}
\toprule
\textbf{Shot} & \multicolumn{2}{c}{\textbf{Non-thinking}} & \multicolumn{2}{c}{\textbf{Thinking}} & \multicolumn{2}{c}{\textbf{NT $-$ T}} \\
              & EMA   & $\kappa$ & EMA   & $\kappa$ & $\Delta$EMA & $\Delta\kappa$ \\
\midrule
0-shot & 0.919 & 0.838 & 0.955 & 0.910 & $-0.036$ & $-0.072$ \\
1-shot & 0.899 & 0.798 & 0.819 & 0.638 & $+0.080$ & $+0.160$ \\
4-shot & 0.952 & 0.905 & 0.886 & 0.773 & $+0.066$ & $+0.132$ \\
\bottomrule
\end{tabular}
\end{table}

\begin{figure}[t]
    \centering
    \includegraphics[width=0.92\linewidth]{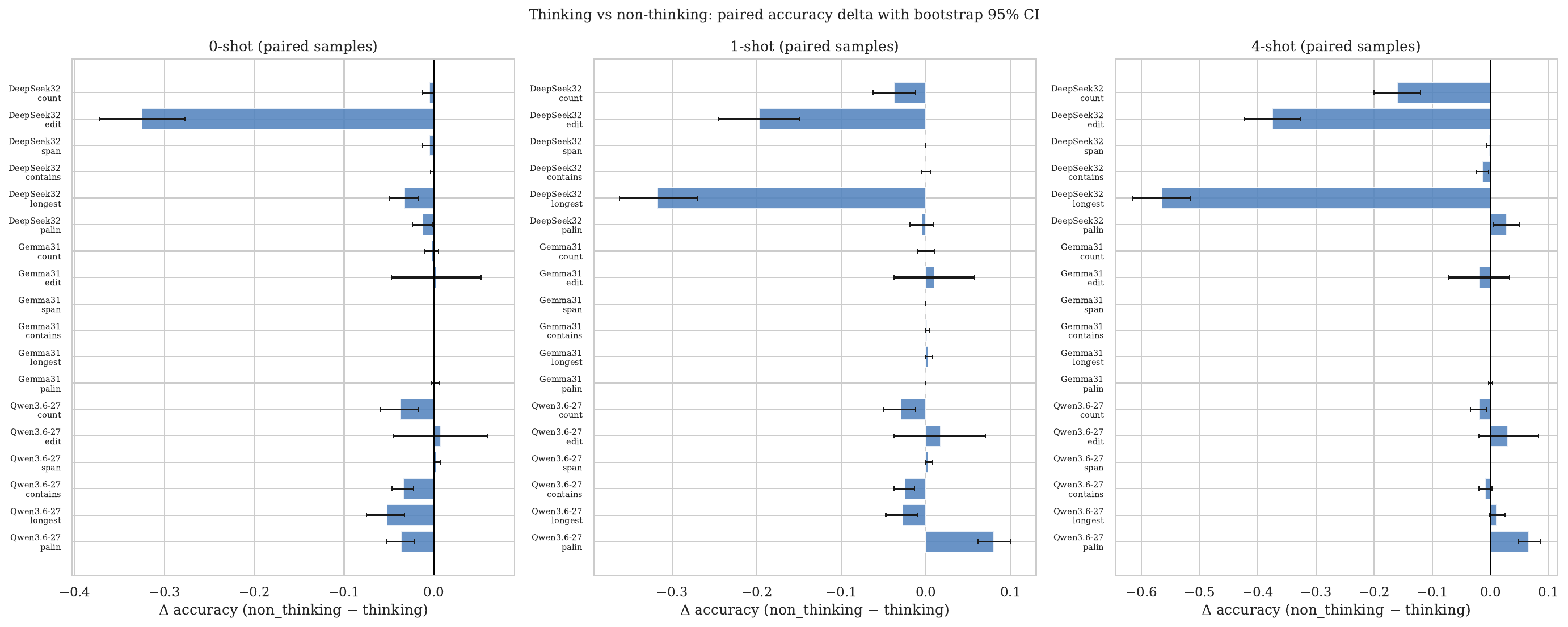}
    \caption{\textbf{Thinking-mode accuracy delta} ($\text{EMA}_\text{non-think} - \text{EMA}_\text{think}$, per task, at 0-, 1-, and 4-shot) with 95\% bootstrap CIs. Bars to the right of zero indicate thinking hurts; bars to the left indicate it helps. Qwen3.6-27B is hurt on \texttt{palindrome\_test}; Gemma4-31B is essentially indistinguishable across modes for character-level tasks at this scale.}
    \label{fig:thinking_delta}
\end{figure}

Three concrete patterns emerge. \emph{(1) Thinking can hurt at fixed scale.} Table~\ref{tab:thinking_palindrome} shows Qwen3.6-27B's non-thinking variant outperforming thinking at 1-shot ($0.899$ vs.\ $0.819$) and 4-shot ($0.952$ vs.\ $0.886$); McNemar's test on the 4-shot combined cell is significant after Bonferroni correction (odds ratio $27.5$). The $\kappa$ drop ($0.905 \to 0.773$) is consistent with the EMA drop and rules out a parsing artifact (PER = 0). \emph{(2) Thinking can be a no-op.} Figure~\ref{fig:thinking_delta} shows Gemma4-31B's thinking and non-thinking results are essentially identical across the four near-saturated tasks at this scale, suggesting that for those character-level operations the model's reasoning trace adds no decision-relevant information. \emph{(3) Thinking can help when reasoning is explicitly trained for.} DeepSeek-R1-32B (thinking-by-design) achieves $99.8\%$ on \texttt{longest\_string\_finder} and $99.4\%$ on \texttt{letter\_container} at 4-shot. The pattern is consistent across English and random conditions, ruling out linguistic-prior explanations. We interpret it as \emph{reasoning-mode interference}: chain-of-thought traces inject surface-format variance that on already-saturated tasks distorts more than it helps. Token-cost and latency penalties of thinking mode are not always offset by accuracy gains for character-level operations; the per-task thinking-cost table (mean tokens, EMA per token) is in Appendix~\ref{app:full_results}, Table~\ref{tab:thinking-cost-efficiency}.

\subsection{Diagnostic Depth: What Aggregate Accuracy Misses}\label{sec:diagnostic}

The framework reports class-conditional metrics, monotonicity over shot count, and effect sizes alongside accuracy. We illustrate two cases where these surface findings that aggregate accuracy hides.

\paragraph{Class-conditional asymmetry on binary tasks.} Qwen3.5-2B on 0-shot \texttt{palindrome\_test} (English) achieves $0.430$ EMA on the positive class (true palindromes) and $0.965$ on the negative class. The model is systematically biased toward ``no'', and aggregate $0.6975$ EMA reads like a noisy classifier. The same disaggregation flags Gemma4-31B's 0-shot \texttt{palindrome\_test} (English) at $\text{TPR}=0.685$ vs.\ $\text{TNR}=1.000$, also a non-trivial under-prediction of the positive class. The examples above are the class-conditional cases reported in the text. A full TPR/TNR/MCC table is not included in the appendix; those per-cell metrics are in the released evaluation CSVs.

\paragraph{Non-monotone response to demonstrations.} Conventional intuition predicts \texttt{monotone\_up}: more examples should help. Kendall's $\tau$ on shot index versus EMA per (model, task, mode, condition) shows this fails systematically for already-saturated tasks. Qwen3.6-27B on \texttt{palindrome\_test} is \texttt{valley\_1shot} in both modes (Table~\ref{tab:thinking_palindrome}): non-thinking $0.919 \to 0.899 \to 0.952$ and thinking $0.955 \to 0.819 \to 0.886$. Kendall's $\tau$ on those combined trajectories is $+0.33$ and $-0.33$. The 1-shot dip is shared; thinking is the mode that stays below its 0-shot accuracy. We propose that monotonicity testing across $\{0, 1, 4\}$ shots becomes a standard diagnostic. The Kendall-$\tau$ sparkline grid for all (model, mode, task, condition) cells is in Appendix~\ref{app:full_results}, Figure~\ref{fig:kendall_tau_grid}.

\subsection{Floor: Positional Substring Extraction}\label{sec:floor}

The five core tasks span saturated to challenging-but-tractable. To probe what current architectures cannot do, we include a sixth task that requires extracting the substring located at a specified character offset within a document of 200--500 words. English items are Wikipedia excerpts (mean 2{,}128 characters). Random items are documents sampled from the random vocabulary (mean 3{,}593 characters): about as many words, but longer words. Across all 11 (model, thinking-mode) configurations and three shot modes, max 4-shot EMA is $\mathbf{6.75\%}$ (GPT-OSS-20B). Gemma4-31B and DeepSeek-R1-32B reach $0.5\%$ and $0.25\%$. Qwen3.5-9B and Qwen3.5-27B reach $0.5\%$. The Qwen3.5-2B 4-shot cell is incomplete: the observed English subset is 0/113, and no random-condition estimate is available. The remaining complete configurations are at exact $0\%$. Scale alone is insufficient. The failure is computational: maintaining a character-position counter across a long document while extracting span contents is beyond every model evaluated. We release this task as a long-horizon discriminator that future architectures can target as the core tasks saturate.

\subsection{Pairwise Model Comparisons}\label{sec:pairwise}

McNemar's test with continuity correction is computed separately within each thinking mode on the 18 combined-condition cells (6 tasks $\times$ 3 shots). Tables~\ref{tab:mcnemar_sig_matrix} and~\ref{tab:mcnemar_sig_matrix_nonthinking} report, for each pair, how many of those cells remain significant after Bonferroni correction across the pair's 18 cells. These tables are not pooled into one ordering: Qwen3.5 was evaluated only without thinking, while GPT-OSS-20B and Nemotron-Cascade-2 were evaluated only with thinking, so several cross-family pairs have no shared-mode test.

In thinking mode, GPT-OSS-20B wins 9 and loses 0 of 18 cells against DeepSeek-R1-32B, wins 12 and loses 0 against Gemma4-31B, wins 12 and loses 0 against Nemotron-Cascade-2, and wins 9 and loses 1 against Qwen3.6-27B. Qwen3.6-27B wins 5, loses 3, and is nonsignificant on 10 cells against Gemma4-31B; it wins 8 and loses 0 against Nemotron. DeepSeek-R1-32B and Qwen3.6-27B each win 2 of their 4 significant cells. In non-thinking mode, Qwen3.5-27B wins 12 and loses 0 against Qwen3.5-9B, and wins 15 and loses 0 against Qwen3.5-2B; Qwen3.5-9B wins 14 and loses 0 against Qwen3.5-2B. Qwen3.5-27B and Qwen3.6-27B are significant on only 3 cells (2 wins for Qwen3.5-27B, 1 for Qwen3.6-27B, 15 nonsignificant). The non-thinking DeepSeek-R1-32B run wins 2 and loses 7 against Qwen3.5-27B, wins 1 and loses 8 against Qwen3.6-27B, and wins 2 and loses 7 against Gemma4-31B. Its comparison with Qwen3.5-9B is mixed, 6 wins each. There is no GPT-OSS-20B versus Qwen3.5 test and no Nemotron versus Qwen3.5 test.

%% file: files/5_discussion.tex
SyntaxBench surfaces three findings whose interpretation depends on the statistical machinery applied: tokenization mediates the English-vs-random behavior usually attributed to lexical familiarity (\S\ref{sec:tokenization}), reasoning mode helps unevenly and sometimes hurts on character-level tasks (\S\ref{sec:thinking}), and positional substring extraction (\texttt{index\_to\_span}) is unsolved across all 11 (model, mode) configurations (\S\ref{sec:floor}).

\paragraph{Tokenization is a confounder, not a covariate.} The standard ``LLMs fail on non-natural inputs'' framing is descriptive: it observes the gap but does not localize the cause. Our chi-square analysis (\S\ref{sec:tokenization}) shows 25 of 197 complete cells exceed the Bonferroni threshold. Random inputs are easier in 23 of those cells and English is easier in 2. The mechanism on the random-easier majority is tokenizer-mediated: random strings pack $1.892$ chars/token, English $3.169$, and \texttt{count\_chars} accuracy declines as English token count rises, while random remains higher in every shared bin. Any character-level evaluation that does not stratify by tokenization risks attributing tokenizer artifacts to model capability.

\paragraph{Reasoning mode is a falsifiable axis.} Paired McNemar across thinking and non-thinking variants of Qwen3.6-27B and Gemma4-31B (\S\ref{sec:thinking}) shows three regimes coexist: thinking can hurt at fixed scale (Qwen3.6 \texttt{palindrome\_test}, $\Delta\text{EMA} = -0.066$ at 4-shot, odds ratio 27.5), be a no-op (Gemma4 across the four near-saturated tasks), or unambiguously help when reasoning is explicitly trained for (DeepSeek-R1). Reporting paired thinking-vs-non-thinking results with token-cost and latency should be standard for any reasoning-toggleable model.

\paragraph{Composite leaderboards are shape-blind.} Macro EMA at 4-shot ranges from 0.65 to 0.82 across the five thinking models. The composite intervals resample the six task-level EMAs, so their width is between-task spread rather than a sample bootstrap: GPT-OSS-20B is $0.818$ [0.512, 0.989]. Averaging over a saturated task ($> 99\%$), a discrimination task (0--89\%), and a floor task ($< 7\%$) hides exactly the structure that distinguishes models. Per-task and per-(mode, shot) reporting is necessary even when a composite is requested.

\paragraph{Recommendations for character-level benchmark design.} (1) Tokenizer stratification by default; the chars-per-token ratio between conditions disentangles familiarity from granularity. (2) Paired statistical tests across modes and shots (McNemar with continuity correction; Kendall's $\tau$ over shot index). (3) Class-conditional reporting on binary tasks (TPR/TNR/MCC/$\kappa$) to surface positive- or negative-class under-prediction. (4) Both Bonferroni and BH-FDR corrections reported alongside; the conservative cutoff anchors headline claims while FDR retains exploratory signal. The remaining headroom on \texttt{edit\_distance} and the floor on \texttt{index\_to\_span} are the two operations most resistant to current mid-sized open-weight transformers; both involve maintaining a counter (over edit operations or character offsets) across a string-valued state and should be implemented with deterministic library calls until architectures designed for long-horizon symbolic state become standard.

%% file: files/6_limitations.tex
Several limitations bound the conclusions of this work.

\textbf{Model and tokenizer coverage.} We evaluate eight open-weight LLMs (2B--32B); frontier-scale models (100B+) and closed-weight models are not included. The tokenizer-stratified analysis (\S\ref{sec:tokenization}) requires public tokenizer access, so closed-tokenizer models are absent from per-bucket results.

\textbf{Closed prompt and language scope.} Each task uses a fixed task-specific system prompt; we do not vary instruction phrasing systematically~\cite{webson2022prompt}. SyntaxBench evaluates English and synthetic random strings only; findings may differ for non-Latin scripts, agglutinative morphology, or scripts with qualitatively different tokenizer behavior.

\textbf{Inference-backend differences.} Models are served via the backend best supported per architecture (vLLM, \texttt{llama.cpp}, Ollama). With deterministic decoding throughout, backend differences in tokenization handling and reasoning-trace stripping may introduce small per-model variations beyond the model itself.

\textbf{Task scope.} The six tasks cover intra-word, inter-word, and positional-substring operations but exclude document-level constrained generation, encoding/decoding, and multi-step algorithmic reasoning, as we deliberately scope to operations with single-pass ground truth and well-defined partial-credit metrics.

\textbf{Statistical and intervention caveats.} We report both Bonferroni and BH-FDR corrections; bootstrap CIs assume sample independence within each cell, which our protocol enforces. The thinking-vs-non-thinking comparison conflates differences in the chat template, decoding length, and (for some models) training data; we treat thinking mode as the deployment-level switch a user has access to, not as a controlled intervention isolating the reasoning trace alone.

%% file: files/8_conclusion.tex
We presented \textbf{SyntaxBench}, a diagnostic benchmark and statistical evaluation framework for character-level reasoning in LLMs. The benchmark contains six deterministic tasks, and the code will be released soon. The five core tasks use paired English and length-matched random-string inputs; the \texttt{index\_to\_span} documents share a 200--500 word band and are not character-length matched. The pipeline reports bootstrap CIs, Cohen's $\kappa$, paired McNemar with odds-ratio effect sizes, Kendall's $\tau$ over shot index, class-conditional binary metrics, tokenizer-stratified accuracy, integer and string error structure, and Bonferroni- and FDR-corrected significance.

Applying this framework to eight open-weight LLMs (2B--32B) across 11 (model, thinking-mode) configurations surfaces three findings that aggregate accuracy hides: \emph{(i)} tokenization mediates the English-vs-random gap, with random strings easier than English in 23 of the 25 Bonferroni-significant cells, contradicting the standard lexical-familiarity framing; \emph{(ii)} reasoning mode is not uniformly beneficial and sometimes hurts at fixed scale (Qwen3.6-27B on \texttt{palindrome\_test} drops 6.6 EMA points at 4-shot in thinking mode); \emph{(iii)} positional substring extraction (\texttt{index\_to\_span}) is unsolved by every model evaluated (max 4-shot EMA 6.75\%), differentiating future architectures as the core tasks saturate.

The methodological contribution generalizes beyond the specific benchmark: character-level LLM evaluation should stratify by tokenization, report paired tests across modes and shots, and treat reasoning-mode and class-conditional behavior as falsifiable hypotheses with explicit effect sizes. The code will be released soon.

%% file: files/appendix/methodology.tex
\section{SyntaxBench Details}
\label{app:syntaxbench}

\subsection{Dataset Construction}
\label{app:data}

\subsubsection{Text Data Collection}
We construct SyntaxBench inputs from three complementary sources. The English vocabulary is drawn from the Natural Language Toolkit (NLTK) corpus (236{,}736 words). For \texttt{palindrome\_test}, we additionally use a curated palindrome word list of 229 entries sourced from Wiktionary\footnote{\url{https://en.wiktionary.org/wiki/Appendix:English_palindromes}}. For \texttt{index\_to\_span}, we use English Wikipedia excerpts and random-vocabulary documents targeted to 200--500 words. Mean length is 2{,}128 characters on the English condition and 3{,}593 characters on the random condition, because the random tokens are longer than Wikipedia's short function words.

\subsubsection{Length-Matched Random-String Corpus}
We generate a synthetic random-string corpus designed to match the empirical character-length distribution of the NLTK English vocabulary. For each English word of length $\ell$, we sample a random string of the same length uniformly from the alphabet $[a\text{-}z]$. This pairing controls for the trivial confound that random strings might differ in length from English words, which would otherwise contaminate any English-vs-random comparison via length effects rather than character-distribution effects. For \texttt{index\_to\_span}, both conditions are targeted to 200--500 words (observed means 321 English and 336 random). Random documents are assembled by sampling that many strings from the random vocabulary, not by replacing each English word in an excerpt. Those strings follow the NLTK word-list lengths rather than Wikipedia's short function words, so mean document length is 3{,}593 characters versus 2{,}128.

The matched-length design for the word-level corpus has been checked against the chi-square analysis: across 197 complete (model, task, shot, mode) comparisons, 25 (12.7\%) show a significant English-vs-random gap after Bonferroni correction. Random inputs are easier in 23 of those cells and English is easier in 2 (\S\ref{sec:tokenization}). One Qwen3.5-2B \texttt{index\_to\_span} cell is incomplete and is excluded from the family.

\subsubsection{Per-Task Sample Counts}
\begin{itemize}
    \item \texttt{count\_chars}: 200 English + 200 random = 400 samples
    \item \texttt{letter\_container}: 400 English + 400 random = 800 samples (balanced across yes/no ground truth)
    \item \texttt{palindrome\_test}: 400 English + 400 random = 800 samples (balanced across yes/no ground truth)
    \item \texttt{edit\_distance}: 200 English pairs + 200 random pairs = 400 samples
    \item \texttt{longest\_string\_finder}: 200 English lists + 200 random lists = 400 samples
    \item \texttt{index\_to\_span}: 200 English documents + 200 random documents = 400 samples
\end{itemize}
Total: 3{,}200 samples per (shot, thinking-mode) configuration. Across the three shot modes and (where applicable) two thinking modes, this gives up to 19{,}200 inference requests per model.

\subsection{Task Definitions}
\label{app:tasks}

We define six syntactic tasks organized along an intra-word vs.\ inter-word structural axis (the five \emph{core} tasks) plus one stress-test task (\texttt{index\_to\_span}). Each task includes (1) a task-specific system prompt, (2) demonstration examples enabling 0-, 1-, and 4-shot evaluation, and (3) test samples with deterministic ground-truth labels.

\subsubsection{Intra-Word Core Tasks}

\textbf{Count Characters (\texttt{count\_chars}).} Given a string $s$, return the integer character length $|s|$. Models must accurately enumerate all characters. The dataset contains 400 samples (200 English words, 200 length-matched random strings). Output type: integer.

\textbf{Letter Container (\texttt{letter\_container}).} Given a string $s$ and a single character $c$, return ``yes'' if $c$ appears in $s$ (case-insensitive), else ``no''. The dataset contains 800 samples balanced across four cells of 200 each: English-yes, English-no, random-yes, random-no. Output type: yes/no.

\textbf{Palindrome Test (\texttt{palindrome\_test}).} Given a string $s$, return ``yes'' if $s$ reads identically forward and backward (case-insensitive), else ``no''. The dataset contains 800 samples balanced across English palindromes, English non-palindromes, random palindromes, and random non-palindromes (200 each). Output type: yes/no.

\subsubsection{Inter-Word Core Tasks}

\textbf{Edit Distance (\texttt{edit\_distance}).} Given a pair of strings $(s_1, s_2)$, return the integer Levenshtein edit distance, the minimum number of single-character insertions, deletions, or substitutions required to transform $s_1$ into $s_2$. The dataset contains 400 samples (200 English word pairs, 200 random pairs). Output type: integer.

\textbf{Longest String Finder (\texttt{longest\_string\_finder}).} Given a list of $\geq 3$ strings with one strictly longest element, return the longest string verbatim (case-insensitive match). The dataset contains 400 samples (200 English-word lists, 200 random-string lists). Output type: string.

\subsubsection{Stress-Test Task}

\textbf{Index to Span (\texttt{index\_to\_span}).} Given a long text document $d$ and two character offsets $(i, j)$, return the substring $d[i:j]$ verbatim. The model must maintain a character-position counter across the document while extracting span contents. The dataset contains 400 samples (200 English Wikipedia excerpts, 200 documents sampled from the random vocabulary), both in the 200--500 word band; mean document length is 2{,}128 characters on the English condition and 3{,}593 characters on the random condition. Output type: string.

\subsection{Prompt Engineering}
We designed prompts to minimize ambiguity and enforce machine-checkable JSON output. Each task uses a fixed task-specific system prompt with multiple natural-language phrasings as variants. All models are instructed to return a JSON object with a single field \texttt{result} holding the answer; the response parser handles whitespace and JSON-formatting variability (\S\ref{sec:scoring}). Initial prompt candidates were generated and manually refined through pilot testing to ensure consistent formatting across model families. The full prompt templates and demonstration pools are released alongside the dataset.

\subsection{Evaluation Protocol Details}
\label{app:eval_protocol}

For each task, models receive the task-specific system prompt followed by 0, 1, or 4 demonstration message pairs (one user turn + one assistant turn per demonstration), then the test query. Demonstrations are drawn from a held-out pool that does not overlap with the test set. Each test instance is executed as an independent inference request that reuses the system prompt and demonstrations but does not include outputs from prior test queries. Requests are dispatched concurrently to the inference backend.

For models with both reasoning configurations (Qwen3.6-27B, Gemma4-31B), each (shot) cell is evaluated twice: once with the chat template's thinking flag enabled (\texttt{enable\_thinking=True}), once disabled. Reasoning content (when emitted in a separate field by reasoning-capable models) is captured for diagnostic purposes (mean thinking-token counts, see Table~\ref{tab:thinking-cost-efficiency}) but excluded from the parsed answer used for scoring. The response parser extracts the value of the \texttt{result} key after stripping whitespace and normalizing JSON. We additionally apply a Markdown-bold (\texttt{**\{...\}**}) recovery rule for one model whose chat template wraps responses in bold markers; this rule is applied uniformly across all models for fairness.

\subsection{Experimental Setup}
\label{app:experimental_setup}

\subsubsection{Hardware Configuration}
\begin{itemize}
    \item \textbf{GPUs:} 2$\times$ NVIDIA L40 (48GB VRAM each)
    \item \textbf{Memory:} 96GB GPU VRAM total, 256GB system RAM
    \item \textbf{Compute Environment:} Ubuntu 24.04 LTS with CUDA 12.4
\end{itemize}

\subsubsection{Models, Modes, and Inference Backends}

We evaluate eight open-weight LLMs across 11 (model, thinking-mode) configurations. Per-(model, mode) backend, quantization, and tensor-parallel-size are summarized below.

\begin{itemize}
    \item \textbf{Qwen3.5-2B / 9B / 27B}~\cite{qwen35}: served via vLLM at BF16, single-GPU (2B/9B) or TP=2 (27B). Non-thinking only.
    \item \textbf{Qwen3.6-27B}~\cite{qwen36}: served via vLLM at FP8, single-GPU. Both thinking and non-thinking modes via the chat template.
    \item \textbf{Gemma4-31B-it}~\cite{gemma4}: served via vLLM at BF16 with tensor-parallel-size 2 (split across both GPUs). Both modes.
    \item \textbf{GPT-OSS-20B}~\cite{gptoss}: served via Ollama. Vendor-native MXFP4 quantization. Thinking-only (Ollama was the only backend with stable support for this model's architecture on Ada-Lovelace GPUs at the time of evaluation).
    \item \textbf{Nemotron-Cascade-2-31B}~\cite{nemotron}: served via \texttt{llama.cpp} with split GGUF shards. Thinking-only.
    \item \textbf{DeepSeek-R1-32B}~\cite{deepseek}: served via vLLM at BF16 with TP=2. Thinking-by-design. Its non-thinking configuration disables reasoning-trace accounting rather than reasoning itself, and is reported separately. Together with the entries above, these are the 11 configurations.
\end{itemize}

All models are instruction-tuned, support system prompts, and produce JSON-formatted output reliably enough to make per-response parsing the primary signal-extraction mechanism rather than format-recovery heuristics.

\subsubsection{Inference Hyperparameters}

All models were evaluated with the following standardized configuration:

\begin{itemize}
    \item \textbf{Temperature:} 0.0 (greedy decoding for reproducibility)
    \item \textbf{Top-p:} 1.0 (nucleus sampling disabled)
    \item \textbf{Max Tokens:} 2048 for non-thinking, 8192 for thinking (sufficient for reasoning traces plus JSON response on all tasks)
    \item \textbf{Repetition Penalty:} 1.0 (disabled)
    \item \textbf{Stop Sequences:} None (rely on EOS token)
\end{itemize}

\subsubsection{Reproducibility}

\begin{itemize}
    \item \textbf{Random Seeds:} Fixed seed (42) for all stochastic operations in the evaluation pipeline (bootstrap, demonstration sampling).
    \item \textbf{Dataset Splits:} No train/test split. Each task is evaluated at 0-, 1-, and 4-shot on a fixed test set.
    \item \textbf{Prompt Templates:} Standardized system prompts and demonstration formats, released alongside the dataset (see \S\ref{app:tasks} and supplementary materials).
    \item \textbf{Deterministic Decoding:} Temperature 0.0 ensures reproducible outputs across runs for a given backend version.
    \item \textbf{Run Manifest:} The code release will include a per-run JSON manifest recording GPU model, backend version, quantization, wall-clock time, and peak VRAM.
\end{itemize}

The evaluation pipeline and the analysis scripts that produce every figure and table in this paper will be released soon. Raw model outputs are stored in JSONL format.

\subsection{Statistical Methods Reference}
\label{app:stats}

This section provides quick reference for the statistical procedures used in the main text.

\textbf{Cohen's $\kappa$}~\cite{cohen1960coefficient}. For binary tasks, we compute unweighted $\kappa$ on the contingency table of (predicted, ground-truth) pairs: $\kappa = (p_o - p_e) / (1 - p_e)$, where $p_o$ is observed agreement and $p_e$ is chance agreement. For integer tasks (\texttt{count\_chars}, \texttt{edit\_distance}), we compute linearly-weighted $\kappa_w$ where the disagreement penalty scales with $|y_\text{pred} - y_\text{true}|$. Interpretation thresholds (Landis-Koch): $\kappa < 0.20$ slight, $0.20$--$0.40$ fair, $0.40$--$0.60$ moderate, $0.60$--$0.80$ substantial, $\geq 0.80$ almost perfect. Negative $\kappa$ indicates systematic anti-agreement.

\textbf{McNemar's test}~\cite{mcnemar1947note}. For a paired-classifier comparison on $n$ samples, let $b_{01}$ be the count of samples where model A is correct and model B incorrect, and $b_{10}$ the reverse. The test statistic with continuity correction is $\chi^2 = (|b_{01} - b_{10}| - 1)^2 / (b_{01} + b_{10})$, distributed as $\chi^2_1$ under the null. The odds ratio $b_{01}/b_{10}$ is reported as effect size; cells with $b_{01} = b_{10} = 0$ are flagged as ceiling effects. The test is also applied per-model in a paired thinking-vs-non-thinking comparison on the same (task, shot) cell.

\textbf{Bootstrap confidence intervals.} 95\% CIs on per-task EMA are computed by resampling samples (not models) with replacement 10{,}000 times, using the percentile method. The composite macro-EMA interval instead resamples the six task-level EMAs, so it measures between-task spread. The random seed is fixed for reproducibility.

\textbf{Chi-square test of independence}~\cite{pearson1900x}. For the English-vs-random comparison on each (model, task, shot, mode) cell, we construct a 2$\times$2 contingency table (condition $\times$ correctness) and apply Pearson's chi-square test with Yates' continuity correction when any cell count is below 5. Cram\'er's $V$ is reported as effect size. We use chi-square rather than McNemar because English and random are independent samples, not paired observations.

\textbf{Kendall's $\tau$}~\cite{kendall1938new} on shot mode. For each (model, task, mode, condition), we compute $\tau$ between shot index $\{0, 1, 4\}$ and EMA. With $n=3$ points, $\tau$ takes values in $\{-1, -0.33, 0.33, 1\}$, and we additionally classify the trajectory pattern by examining the actual values: \texttt{monotone\_up} ($\tau=1$), \texttt{monotone\_down} ($\tau=-1$), \texttt{peak\_1shot} (max at shot 1), \texttt{valley\_1shot} (min at shot 1), or \texttt{flat} (no variation).

\textbf{Multiple-comparison correction.} We apply both \emph{Bonferroni} and \emph{Benjamini-Hochberg FDR}~\cite{benjamini1995controlling} corrections at $\alpha = 0.05$ over the relevant comparison family. Headline tables in the main text use the more conservative Bonferroni cutoff; the FDR-corrected $q$-values are released in the supplementary CSVs (\texttt{evaluation/mcnemar\_*.csv}, \texttt{evaluation/chi2\_eng\_vs\_rand.csv}). The McNemar matrix is corrected over all model pairs in each (task, shot, mode, condition) cell; the chi-square English-vs-random tests are corrected over the 197 comparisons that contain both conditions. The one incomplete cell is excluded, and Benjamini--Hochberg FDR is applied to the same 197-comparison family.

\subsection{Tokenizer-Stratified Analysis Methods}
\label{app:tokenization}

For each model with a publicly-available HuggingFace tokenizer, we tokenize the raw input content (the text inside the \texttt{<string>}, \texttt{<list>}, etc., XML-style markers used by the system prompts) and report (i) per-(model, task, condition) mean tokens, mean chars-per-token, and per-bin sample counts, and (ii) per-(model, task, condition, token-bin) accuracy at 0-shot. The tokenizer used per model is recorded in \texttt{analysis/tokenization/tokenization\_full.csv}; for closed-tokenizer models we omit the model from the tokenizer-stratified analysis but retain it in the chi-square English-vs-random analysis (which does not depend on tokenizer access).

Cohen's $d$ is computed on per-sample tokens per character, pooled across the five core tasks and all public tokenizers. The single-character \texttt{letter} query tag is excluded, as is \texttt{index\_to\_span}. With $n = 8{,}000$ samples in each condition, the means are $0.542$ (random) and $0.344$ (English) tokens per character, and $d = (\bar x_\text{random} - \bar x_\text{English}) / s_\text{pooled} = 2.165$. The corresponding mean characters per token are $1.892$ and $3.169$. The inputs and inclusion rule are recorded in \texttt{analysis/tokenization/tokenization\_headline.csv}.

\subsection{Datasheet Summary}
\label{app:datasheet}

\textbf{Motivation.} SyntaxBench was created to evaluate LLM character-level reasoning under deterministic ground truth, with paired English and random inputs and statistical machinery sufficient to surface tokenization, reasoning-mode, and class-bias effects. Length matching applies to the five core tasks.

\textbf{Composition.} Six tasks, 3{,}200 samples per complete (shot, thinking-mode) configuration, with English and random conditions. The Qwen3.5-2B \texttt{index\_to\_span} 4-shot cell is incomplete (\S\ref{sec:floor}). Inputs are non-personal lexical items (words, character lists, public Wikipedia excerpts). No personally-identifiable information is included.

\textbf{Collection process.} English vocabulary is the NLTK corpus; palindromes are sourced from Wiktionary; Wikipedia excerpts are used under CC~BY-SA. Random strings are programmatically generated (seeded). The construction scripts will be released soon.

\textbf{Preprocessing.} For the five core tasks, length-matching pairs each English word of length $\ell$ with a uniformly-random string of length $\ell$. \texttt{index\_to\_span} random documents are sampled from this vocabulary into the same 200--500 word band and are not matched to the English character lengths. Demonstrations are sampled from a held-out pool with fixed seed.

\textbf{Uses.} The dataset is intended for diagnostic evaluation of character-level reasoning, comparison of tokenizer-mediated effects, and analysis of reasoning-mode behavior. It is not intended for safety, fairness, or capability claims about LLMs as a whole.

\textbf{Distribution.} The dataset under \texttt{data/} is released under CC~BY-SA~4.0, the same share-alike terms required by the Wikipedia excerpts. The code will be released soon. The Wiktionary-derived palindrome word list retains the attribution and share-alike terms of its CC~BY-SA~3.0 source. A machine-readable Croissant metadata record, \texttt{croissant.json}, accompanies the dataset.

\textbf{Maintenance.} The code will be released soon. Updates will use semantic-version bumps. The release manifest records all evaluated model versions and inference backends.

%% file: files/appendix/tasks.tex
\section{Supplementary Tables and Figures}
\label{app:full_results}

This appendix contains supplementary tables and figures referenced from the main text. All tables are generated by the analysis code, which will be released soon. Readers can reproduce them from \texttt{evaluation/metrics\_summary.csv} and the per-test CSVs. Per-sample model outputs are stored in JSONL format, one file per (model, task, shot, mode) cell.

\subsection{Aggregate Performance Tables}
\label{app:tables_main}

This subsection collects the full aggregate tables: per-task EMA in thinking mode (Table~\ref{tab:main_results}) and non-thinking mode (Table~\ref{tab:main_results_nonthinking}), the shot ablation (Table~\ref{tab:shot_ablation}), Cohen's $\kappa$ (Table~\ref{tab:kappa}), the pairwise McNemar significance matrices for both modes (Tables~\ref{tab:mcnemar_sig_matrix} and~\ref{tab:mcnemar_sig_matrix_nonthinking}), Kendall's $\tau$ over shot count (Table~\ref{tab:kendall_tau}), the English-vs-random tests (Table~\ref{tab:eng_vs_rand}), parse-error rates (Table~\ref{tab:parse_errors}), delta distributions for the integer tasks (Tables~\ref{tab:delta_dist_count_chars} and~\ref{tab:delta_dist_edit_distance}), thinking-cost accounting (Table~\ref{tab:thinking-cost-efficiency}), effect sizes (Table~\ref{tab:effect-sizes}), and the full composite leaderboards (Tables~\ref{tab:composite_leaderboard_full} and~\ref{tab:composite_leaderboard_nonthinking}).

\input{tables/table_main_thinking}

\input{tables/table_main_nonthinking}

\input{tables/table_shot_ablation}

\input{tables/table_kappa}

\input{tables/table_significance_matrix}

\input{tables/table_significance_matrix_nonthinking}

\input{tables/table_kendall_tau}

\input{tables/table_eng_vs_rand}

\input{tables/table_parse_errors}

\input{tables/table_delta_dist}

\input{tables/table_thinking_cost}

\input{tables/table_effect_sizes}

\input{tables/table_composite_leaderboard}

\input{tables/table_composite_leaderboard_nonthinking}

\FloatBarrier

\subsection{Supplementary Figures}
\label{app:figures}

\begin{figure}[!htbp]
    \centering
    \includegraphics[width=0.95\linewidth]{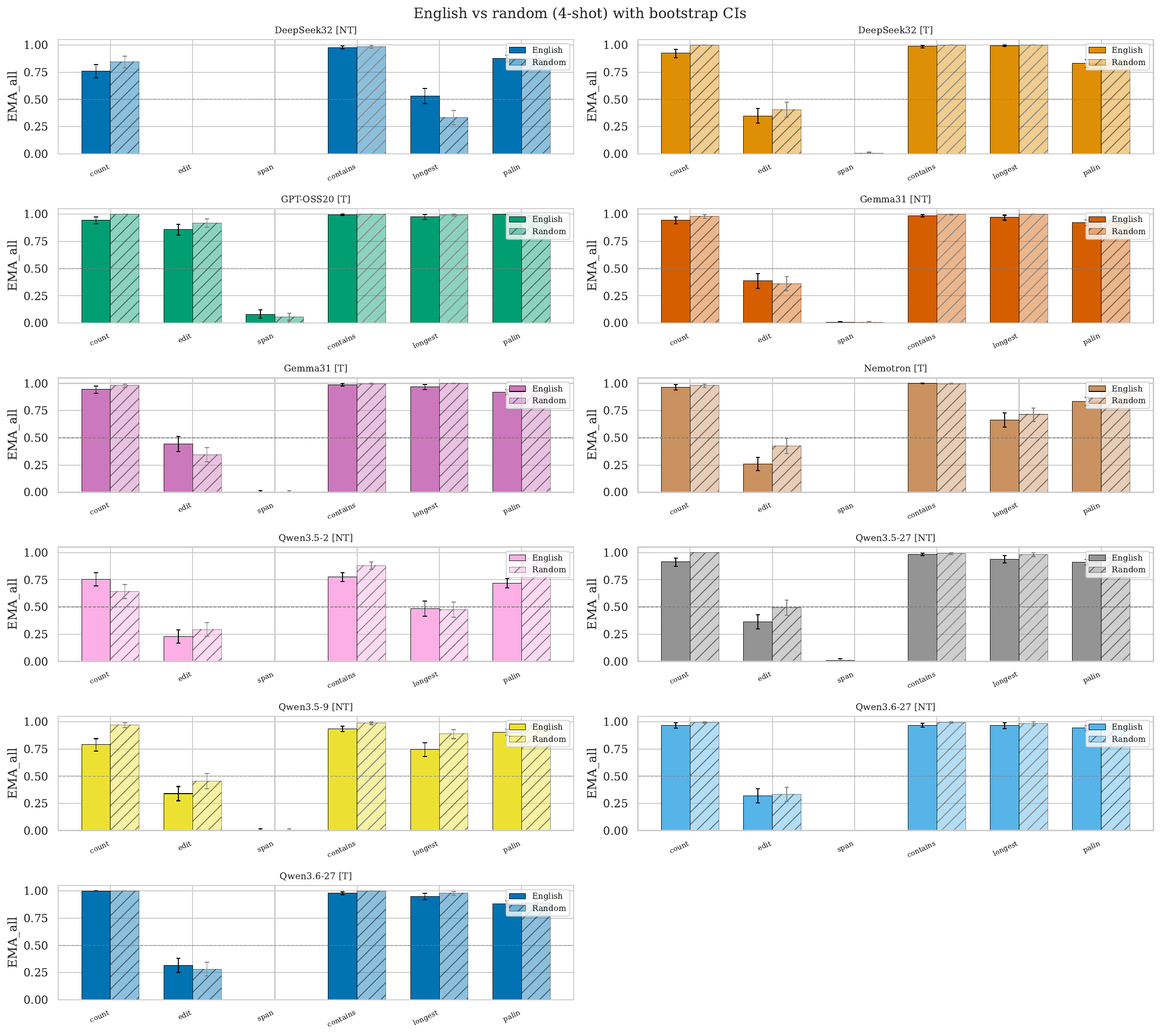}
    \caption{\textbf{English vs.\ random EMA gap} per (model, mode, task) at 4-shot, with 95\% bootstrap CIs. After Bonferroni correction over 197 complete comparisons, 25 cells (12.7\%) show a significant gap; random inputs are easier in 23 of them (\S\ref{sec:tokenization}).}
    \label{fig:eng_vs_rand_gap}
\end{figure}

\begin{figure}[!htbp]
    \centering
    \includegraphics[width=0.95\linewidth]{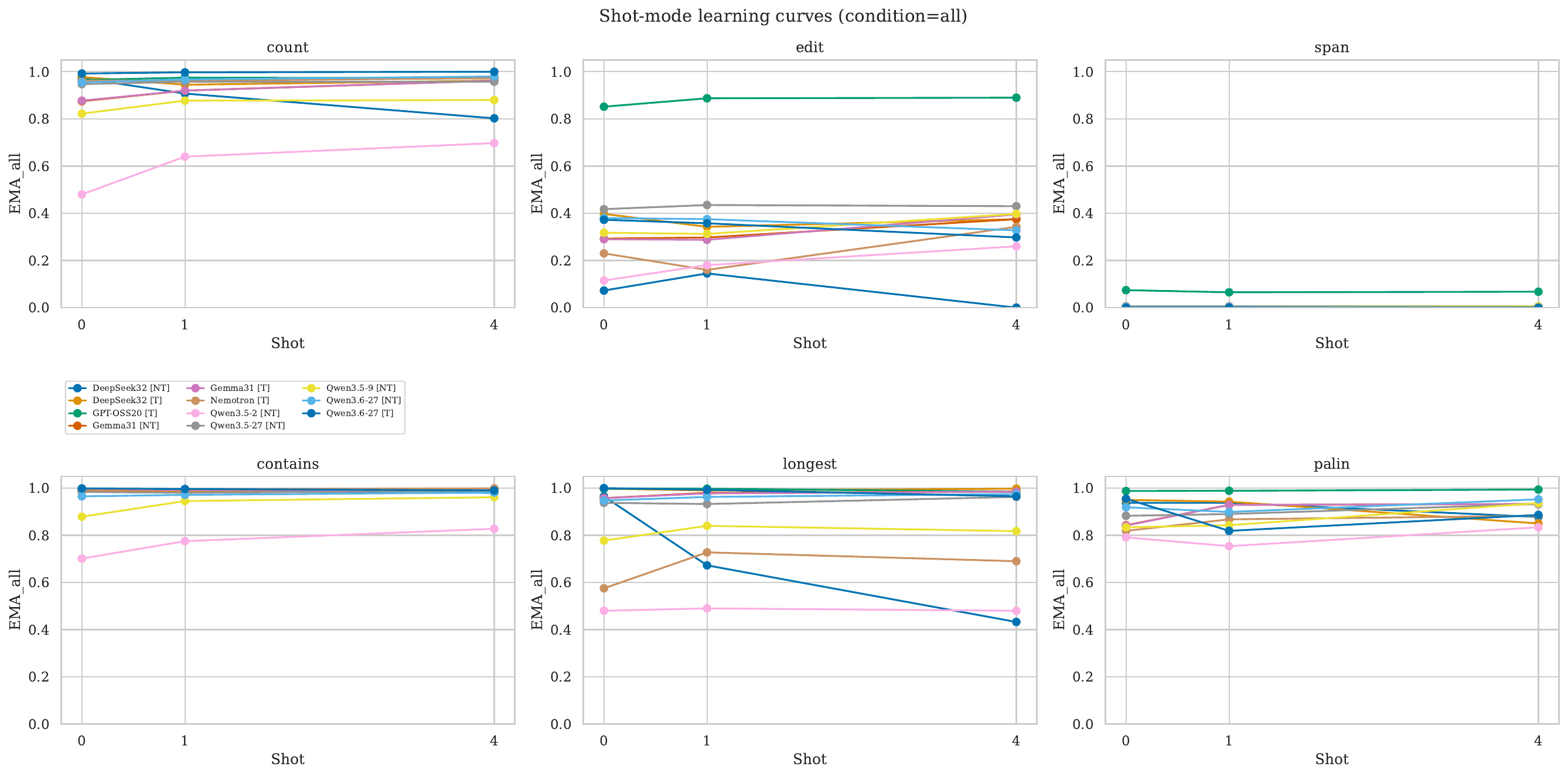}
    \caption{\textbf{Shot-mode learning curves} per (model, mode, task, condition). Reading horizontally shows the per-(model, mode) trajectory across 0/1/4-shot; descending curves correspond to the few-shot interference effect on saturated tasks (\S\ref{sec:diagnostic}).}
    \label{fig:shot_learning_curves}
\end{figure}

\begin{figure}[!htbp]
    \centering
    \includegraphics[width=0.85\linewidth]{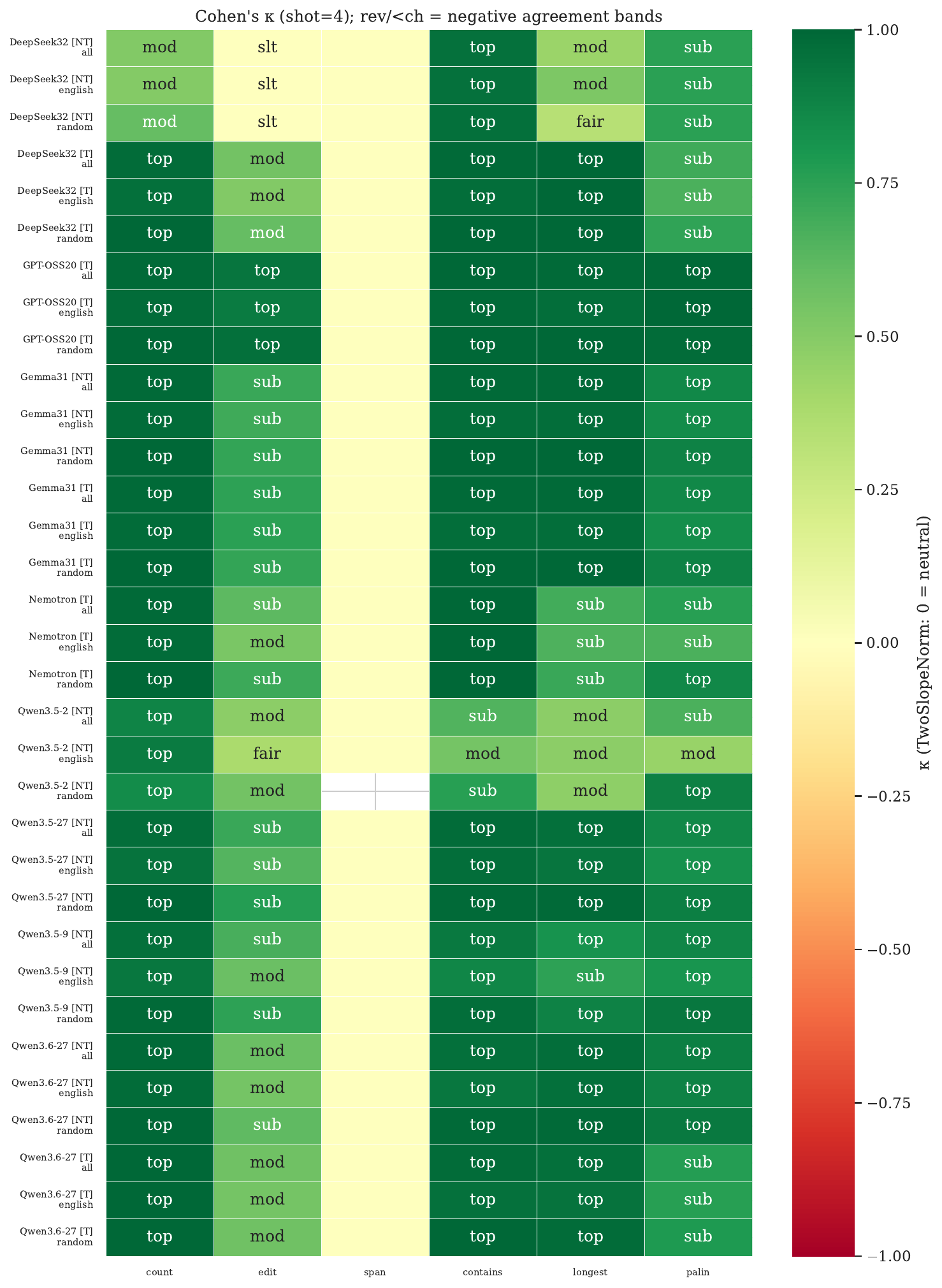}
    \caption{\textbf{Cohen's $\kappa$ heatmap} per (model, mode, task, shot). Cells with $\kappa < 0$ indicate systematic anti-agreement (rare in the post-parser-fix data). The thinking-mode degradation on Qwen3.6 \texttt{palindrome\_test} (\S\ref{sec:thinking}) is visible as the lower $\kappa$ in the thinking row vs.\ the non-thinking row at 1- and 4-shot.}
    \label{fig:kappa_heatmap}
\end{figure}

\begin{figure}[!htbp]
    \centering
    \includegraphics[width=0.95\linewidth]{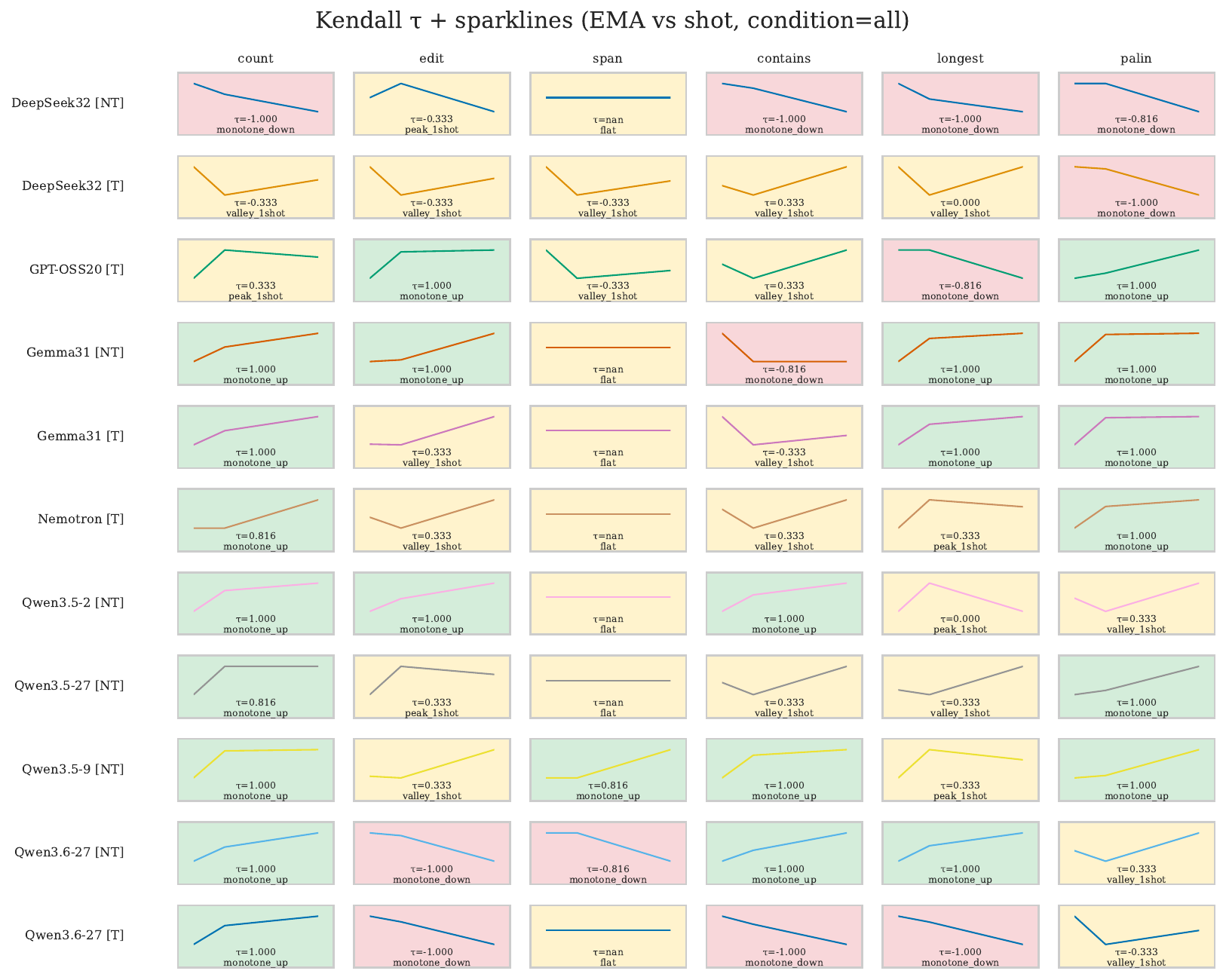}
    \caption{\textbf{Kendall-$\tau$ sparkline grid} over shot mode per (model, mode, task, condition). Trajectory classification (\texttt{monotone\_up}, \texttt{monotone\_down}, \texttt{peak\_1shot}, \texttt{valley\_1shot}, \texttt{flat}) is shown by the cell color and the embedded $\tau$ value. Sign-flips between thinking and non-thinking variants of the same model on the same task are the main signal underlying \S\ref{sec:diagnostic}.}
    \label{fig:kendall_tau_grid}
\end{figure}

\begin{figure}[!htbp]
    \centering
    \includegraphics[width=0.85\linewidth]{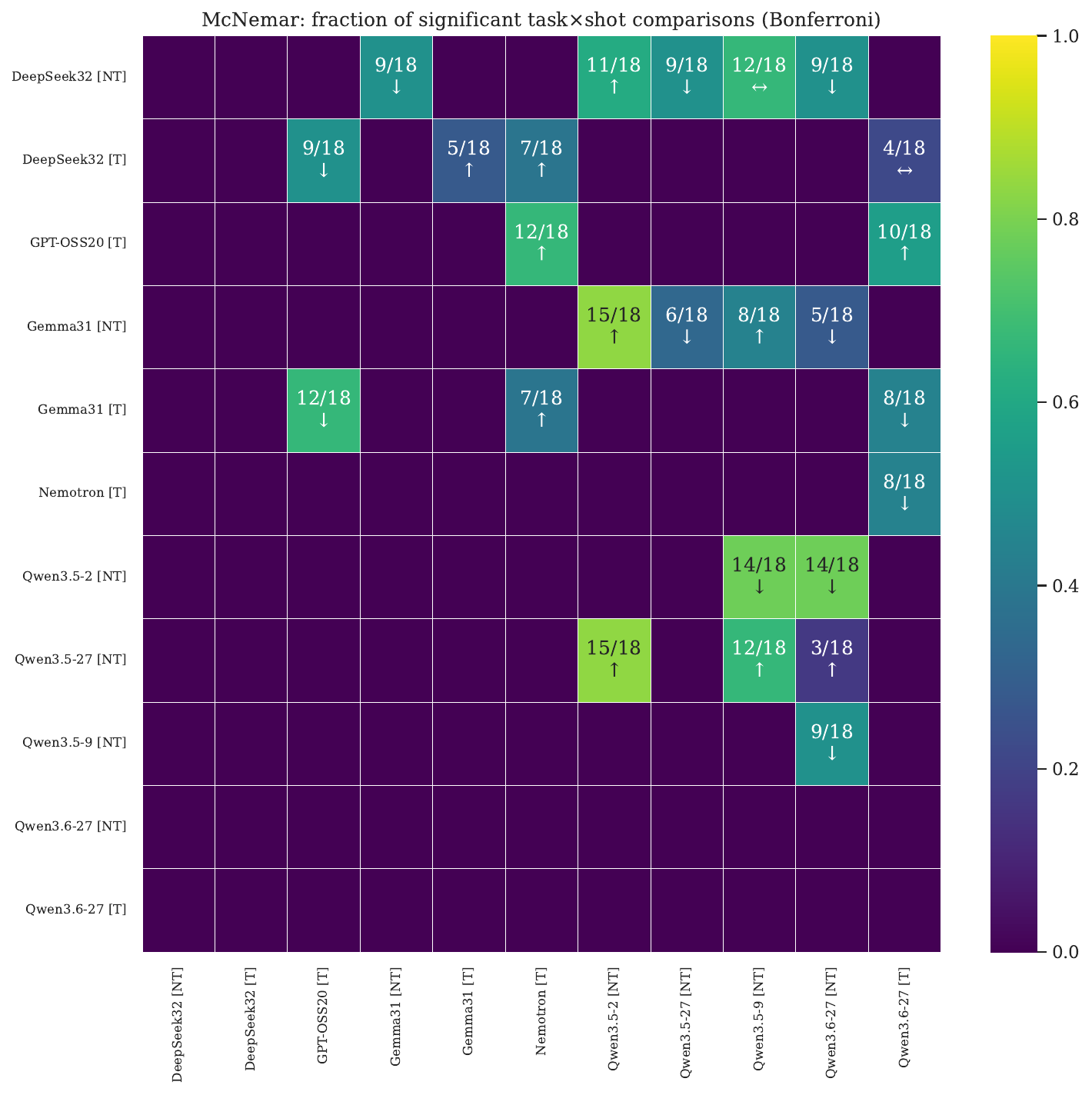}
    \caption{\textbf{McNemar pairwise significance summary.} For each ordered pair of models, the proportion of combined-condition task$\times$shot cells where the row model significantly outperforms the column model after Bonferroni correction within that pair. Thinking and non-thinking pairs are summarized separately in \S\ref{sec:pairwise}.}
    \label{fig:mcnemar_summary}
\end{figure}

\begin{figure}[!htbp]
    \centering
    \includegraphics[width=0.95\linewidth]{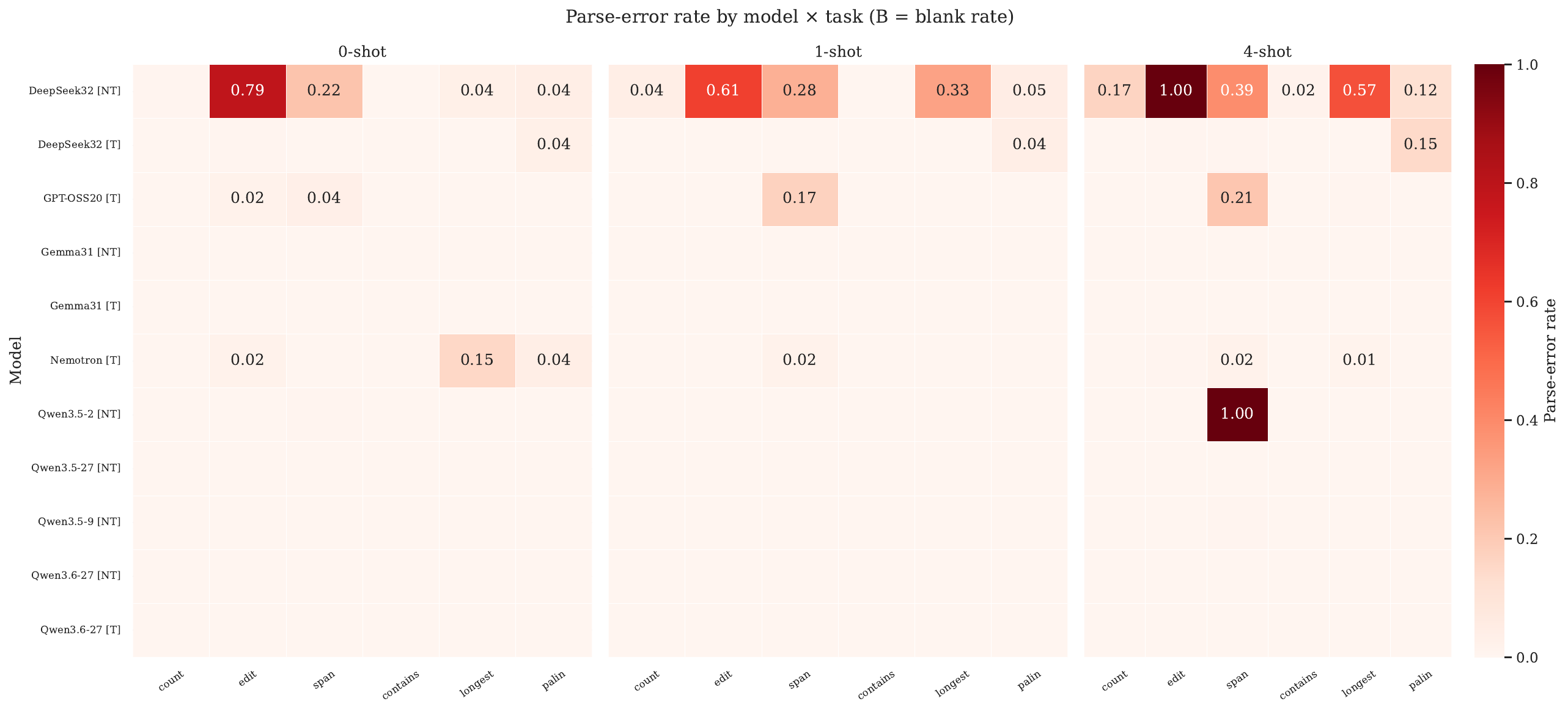}
    \caption{\textbf{Parse error rates} by (model, mode, task, shot). Most models have PER $\leq 0.04$ at 4-shot in any mode; cells with elevated PER appear in the appendix table~\ref{tab:parse_errors}.}
    \label{fig:parse_errors}
\end{figure}

\begin{figure}[!htbp]
    \centering
    \includegraphics[width=0.95\linewidth]{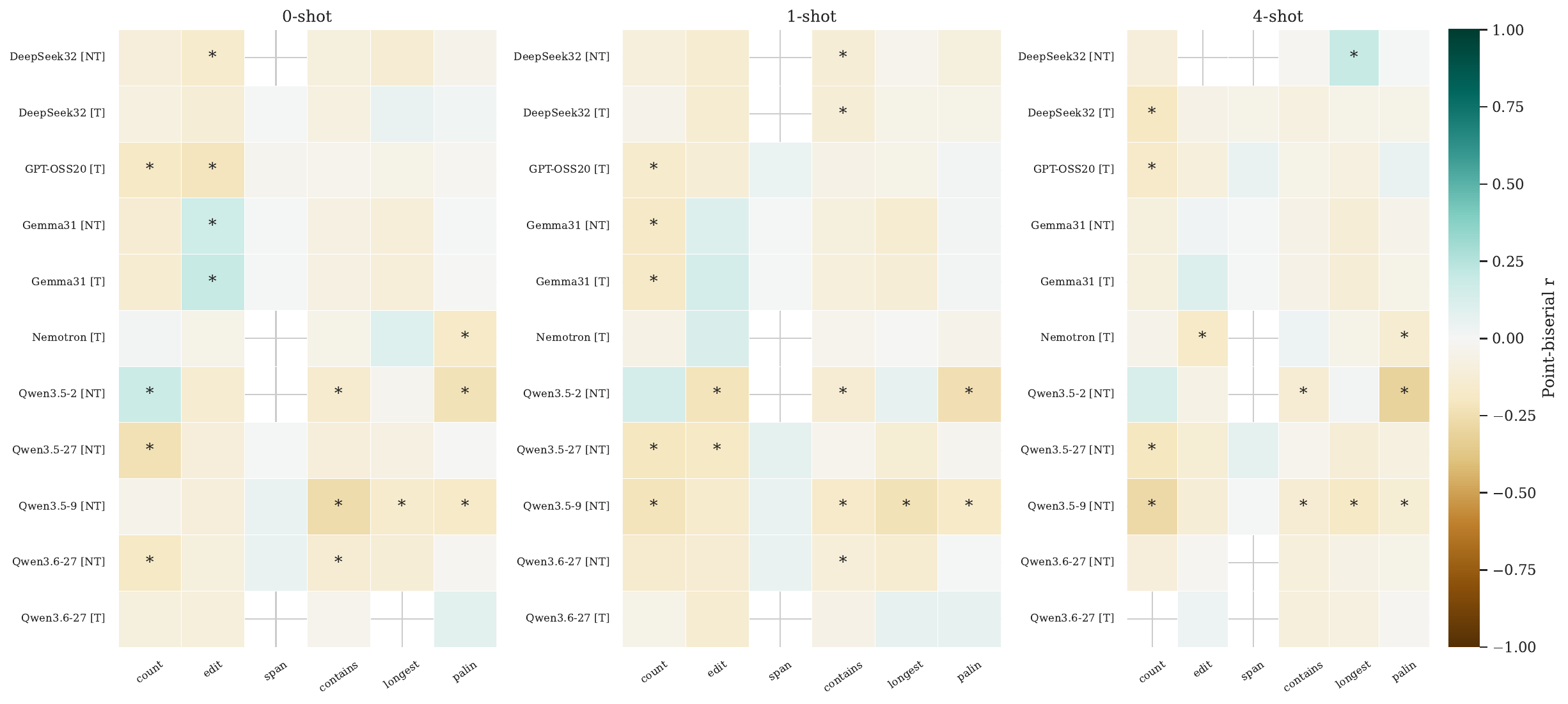}
    \caption{\textbf{Point-biserial correlation between input condition and correctness.} Cells indicate the strength and direction of the English-vs-random effect per (model, mode, task, shot). Magnitudes are uniformly small ($|r| < 0.30$). On the cells that survive Bonferroni correction, random is easier in 23 of 25 (\S\ref{sec:tokenization}).}
    \label{fig:point_biserial}
\end{figure}

\begin{figure}[!htbp]
    \centering
    \includegraphics[width=0.95\linewidth]{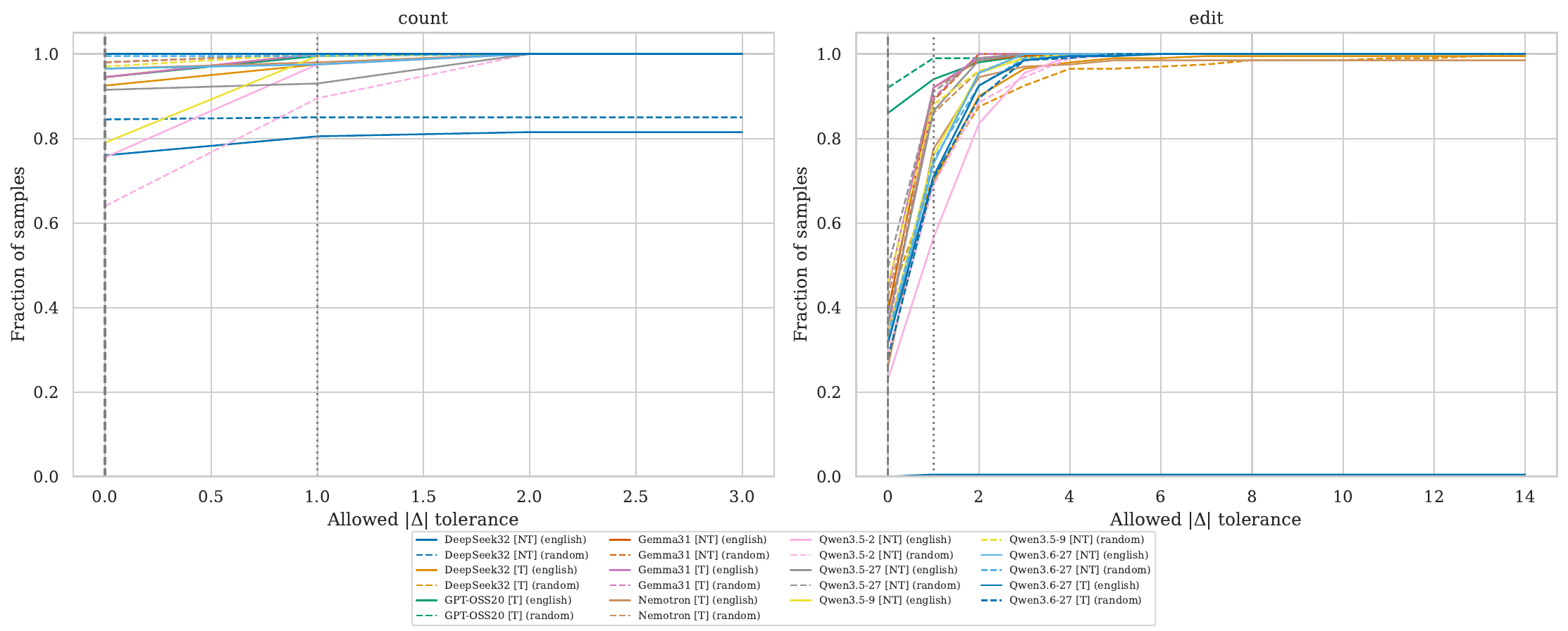}
    \caption{\textbf{Cumulative accuracy as a function of allowed delta tolerance} on integer-output tasks at 4-shot. The gap between $x=0$ (exact match) and $x=1$ (within $\pm 1$) is the off-by-one error rate. For \texttt{edit\_distance}, this gap is large for all models except GPT-OSS-20B, indicating models reason \emph{near} the correct edit distance but rarely \emph{at} it.}
    \label{fig:cumulative_accuracy}
\end{figure}

\begin{figure}[!htbp]
    \centering
    \includegraphics[height=0.7\textheight,keepaspectratio]{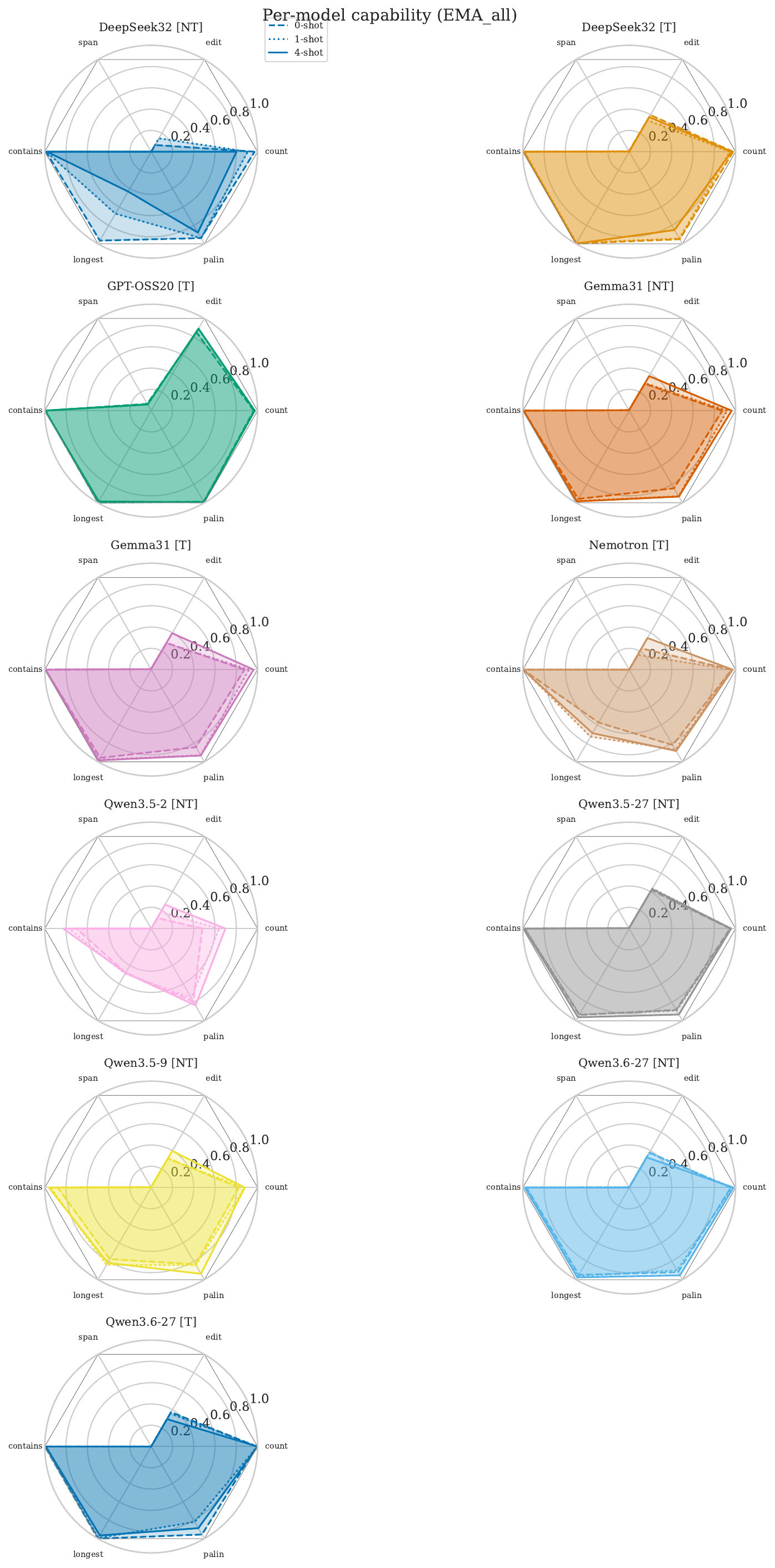}
    \caption{\textbf{Per-(model, mode) capability profile.} Radar chart showing 4-shot EMA on each task. The asymmetry between the four near-saturated tasks and \texttt{edit\_distance} / \texttt{index\_to\_span} (the discrimination and floor tasks) is visible as the radial spread.}
    \label{fig:radar}
\end{figure}

\begin{figure}[!htbp]
    \centering
    \includegraphics[width=0.95\linewidth]{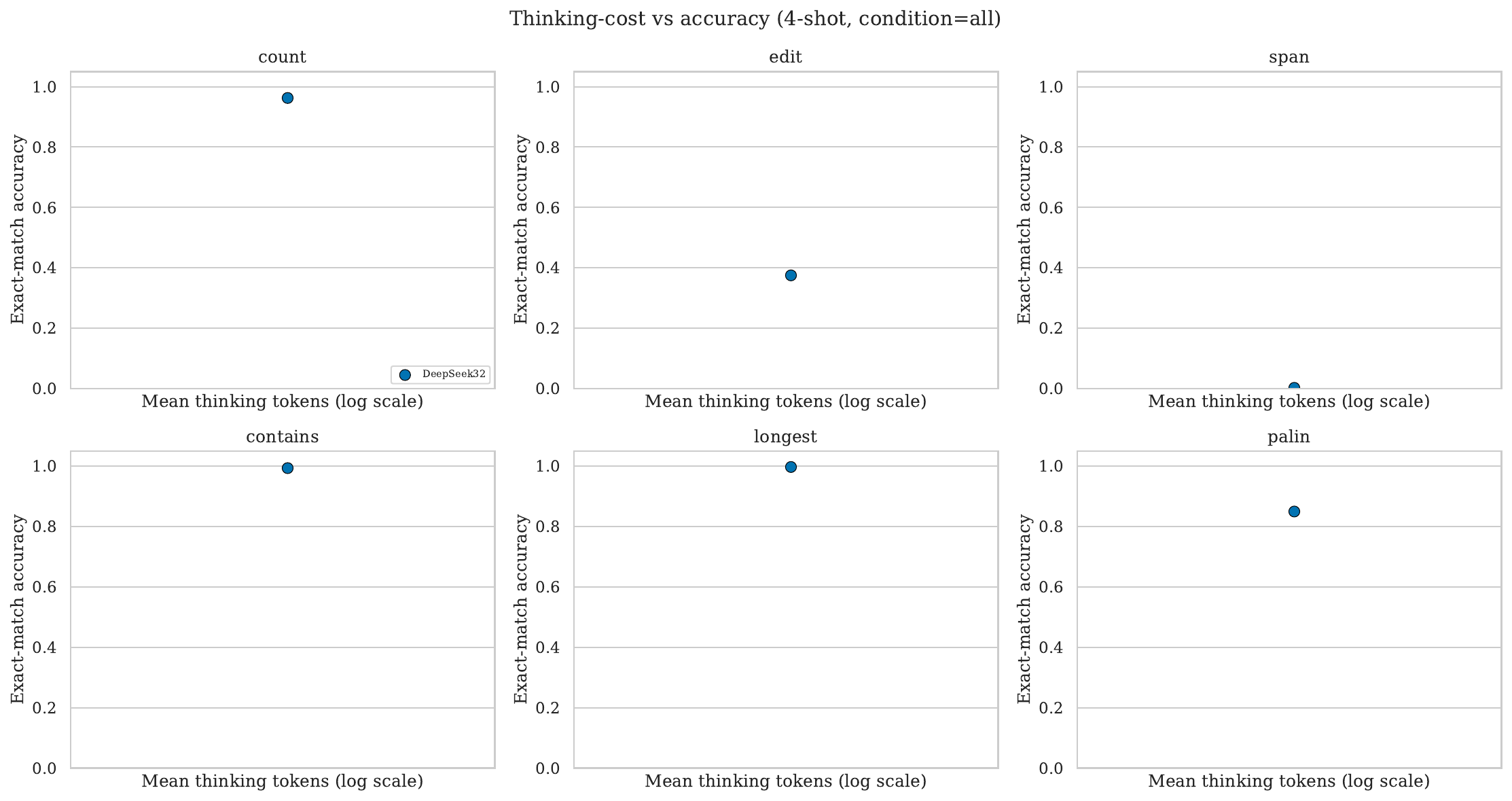}
    \caption{\textbf{Thinking-cost Pareto frontier.} EMA versus mean thinking-token count per (model, task) at 4-shot in thinking mode. Models above-and-left of the Pareto curve are accuracy-efficient; the implications for deployment cost trade-offs are discussed in \S\ref{sec:thinking}.}
    \label{fig:thinking_cost_pareto}
\end{figure}

\begin{figure}[!htbp]
    \centering
    \includegraphics[width=0.95\linewidth]{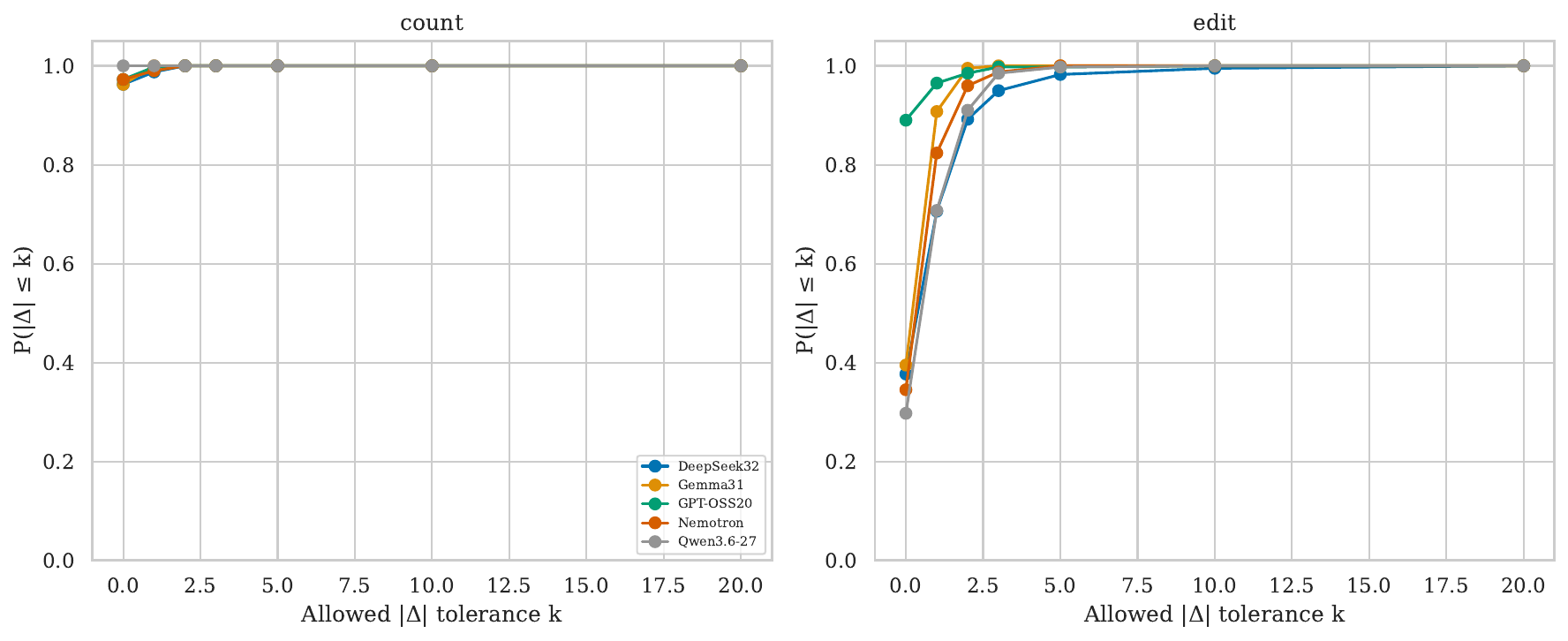}
    \caption{\textbf{Tolerance curves} for integer tasks. Reading from $x=0$ (exact match) to $x=k$ (within $\pm k$) shows the cumulative accuracy as the tolerance widens. The shape distinguishes models whose errors are off-by-one from those whose errors are arbitrary.}
    \label{fig:tolerance_curves}
\end{figure}

\begin{figure}[!htbp]
    \centering
    \includegraphics[width=0.85\linewidth]{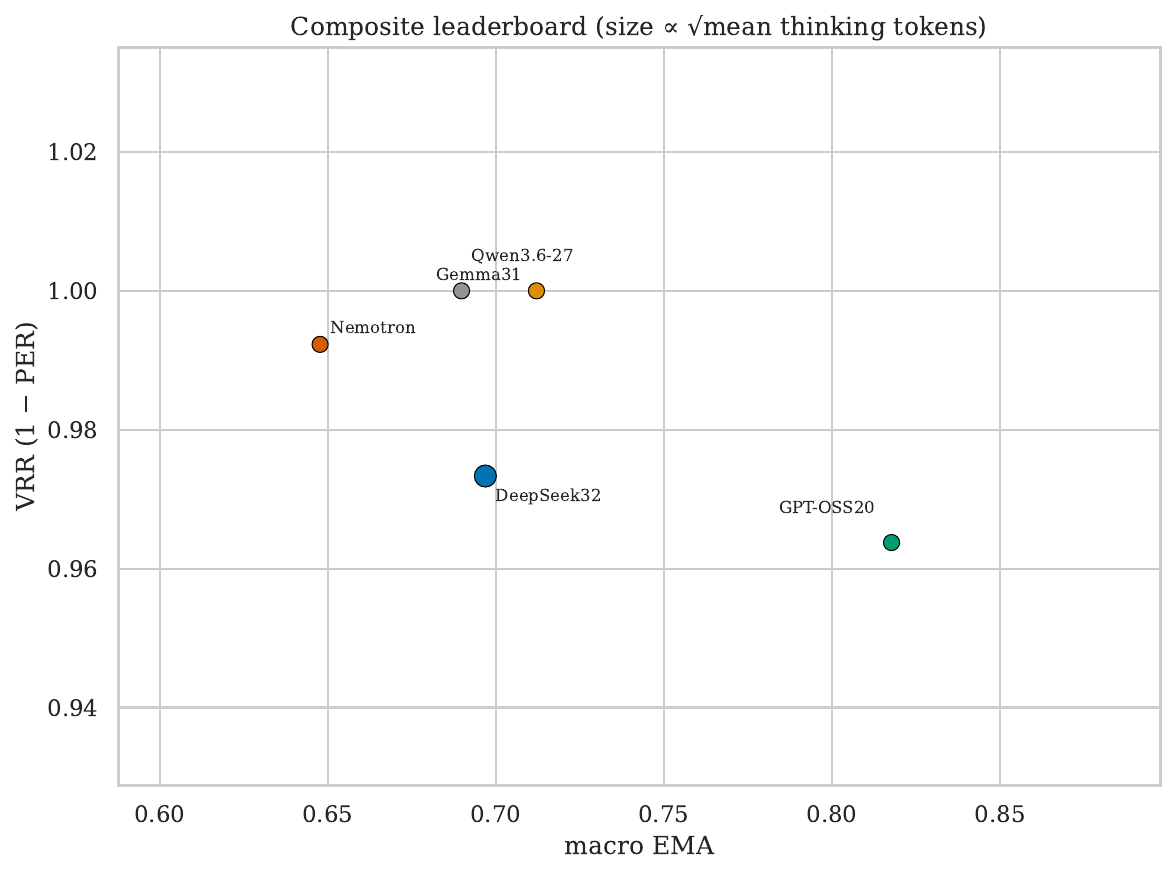}
    \caption{\textbf{Composite leaderboard Pareto frontier.} Macro EMA versus VRR (valid-response rate) per (model, mode) at 4-shot. The composite leaderboard in \S\ref{sec:setup}, Table~\ref{tab:composite_leaderboard}, ranks models on Macro EMA; this figure visualizes the alternative trade-off where parse-error-tolerant models can be preferred under a deployment budget.}
    \label{fig:composite_pareto}
\end{figure}

\begin{figure}[!htbp]
    \centering
    \includegraphics[width=0.85\linewidth]{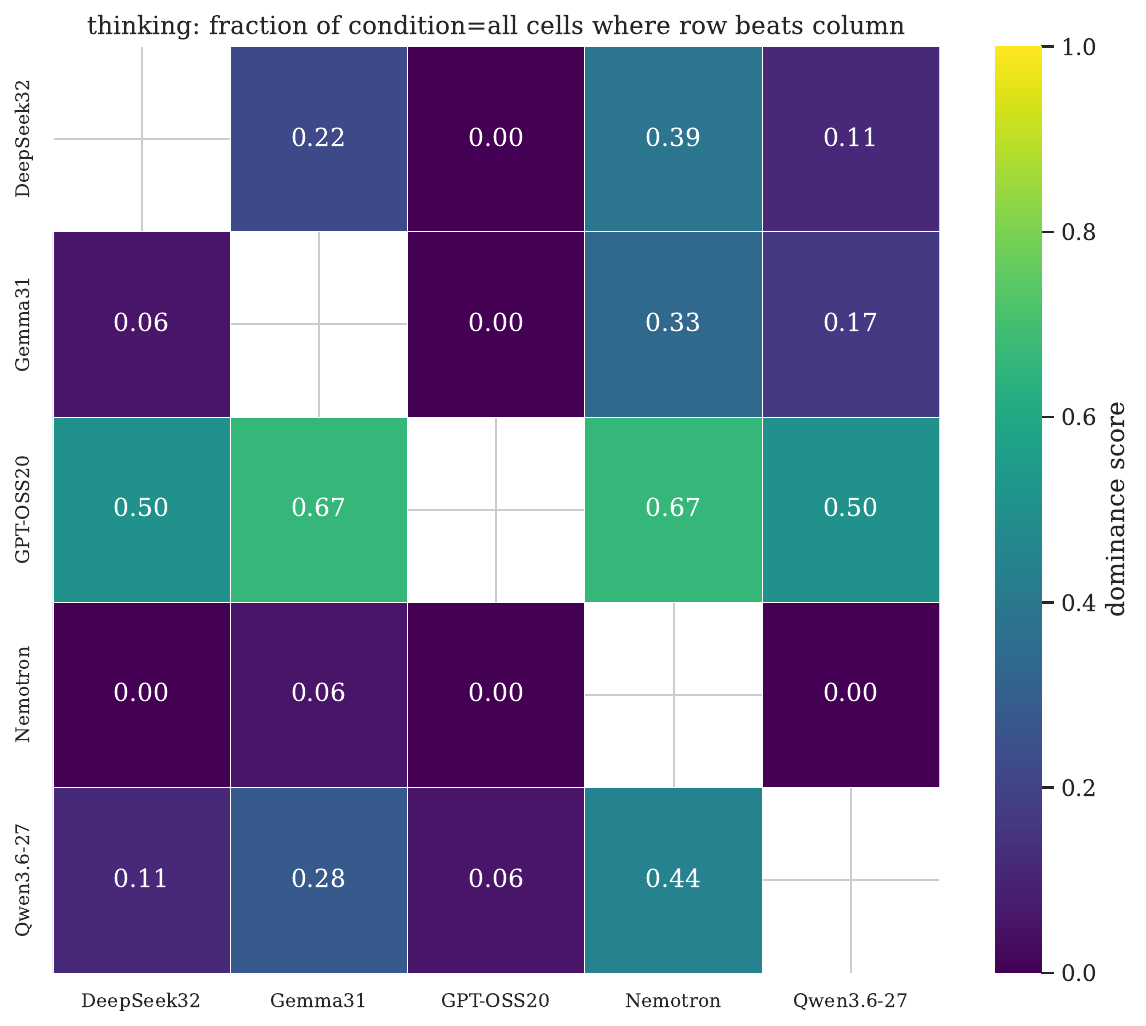}
    \caption{\textbf{Thinking-mode pairwise dominance.} Each cell is the fraction of combined-condition task$\times$shot cells on which the row model significantly beats the column model under the per-cell Bonferroni correction. Thinking and non-thinking comparisons are not pooled, and this matrix does not define a cross-mode ordering.}
    \label{fig:dominance_graph}
\end{figure}

\begin{figure}[!htbp]
    \centering
    \includegraphics[width=0.85\linewidth]{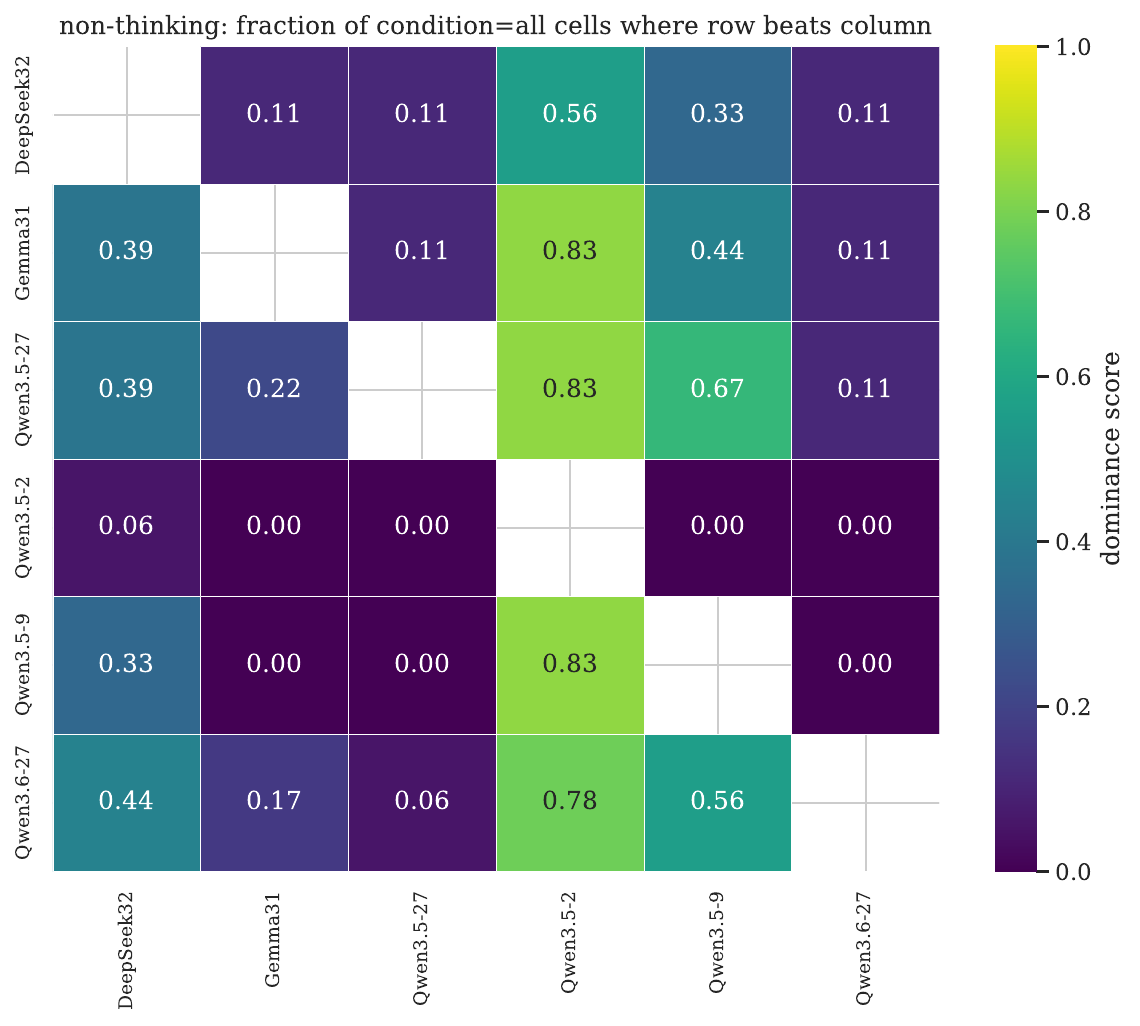}
    \caption{\textbf{Non-thinking pairwise dominance.} Same definition as Figure~\ref{fig:dominance_graph}, restricted to non-thinking runs. Qwen3.5 appears only in this figure; GPT-OSS-20B and Nemotron-Cascade-2 appear only in the thinking-mode figure.}
    \label{fig:dominance_graph_nonthinking}
\end{figure}

The supplementary figures visualize the English-vs-random gap (Figure~\ref{fig:eng_vs_rand_gap}), shot-mode learning curves (Figure~\ref{fig:shot_learning_curves}), Cohen's $\kappa$ (Figure~\ref{fig:kappa_heatmap}), Kendall-$\tau$ trajectories (Figure~\ref{fig:kendall_tau_grid}), the McNemar pairwise summary (Figure~\ref{fig:mcnemar_summary}), parse-error rates (Figure~\ref{fig:parse_errors}), point-biserial condition effects (Figure~\ref{fig:point_biserial}), tolerance behavior on integer tasks (Figures~\ref{fig:cumulative_accuracy} and~\ref{fig:tolerance_curves}), per-model capability profiles (Figure~\ref{fig:radar}), thinking-cost trade-offs (Figure~\ref{fig:thinking_cost_pareto}), and the composite Pareto frontier and mode-specific dominance graphs (Figures~\ref{fig:composite_pareto},~\ref{fig:dominance_graph}, and~\ref{fig:dominance_graph_nonthinking}).

\FloatBarrier

%% file: tables/table_main_thinking.tex
\begin{table}[t]
\centering
\caption{Main results (thinking mode where supported): Exact Match Accuracy (EMA, \%) at 4-shot with 95\% bootstrap CI.
Significance vs.\ best model: $^{*}p{<}0.05$, $^{**}p{<}0.01$, $^{***}p{<}0.001$
(Bonferroni-corrected McNemar's test).}
\label{tab:main_results}
\resizebox{\textwidth}{!}{%
\begin{tabular}{lcccccccccccc}
\toprule
\multicolumn{2}{c}{count\_chars} & \multicolumn{2}{c}{edit\_distance} & \multicolumn{2}{c}{index\_to\_span} & \multicolumn{2}{c}{letter\_container} & \multicolumn{2}{c}{longest\_string\_finder} & \multicolumn{2}{c}{palindrome\_test} \\
\cmidrule(lr){2-3} \cmidrule(lr){4-5} \cmidrule(lr){6-7} \cmidrule(lr){8-9} \cmidrule(lr){10-11} \cmidrule(lr){12-13} & english & random & english & random & english & random & english & random & english & random & english & random \\
\midrule
deepseek\_r1\_32b
 & $92.5_{88.5,96.0}***$
 & $\mathbf{100.0}_{100.0,100.0}$
 & $34.5_{28.0,41.5}***$
 & $40.5_{33.5,47.5}***$
 & $0.0_{0.0,0.0}***$
 & $0.5_{0.0,1.5}*$
 & $98.8_{97.5,99.8}$
 & $\mathbf{100.0}_{100.0,100.0}$
 & $\mathbf{99.5}_{98.5,100.0}$
 & $\mathbf{100.0}_{100.0,100.0}$
 & $83.2_{79.5,86.8}***$
 & $86.8_{83.2,90.0}***$
 \\
gemma4\_31b
 & $94.5_{91.0,97.5}**$
 & $98.0_{96.0,99.5}$
 & $44.5_{37.5,51.5}***$
 & $34.5_{28.0,41.0}***$
 & $0.5_{0.0,1.5}**$
 & $0.5_{0.0,1.5}*$
 & $98.8_{97.5,99.8}$
 & $99.8_{99.2,100.0}$
 & $97.0_{94.5,99.0}$
 & $\mathbf{100.0}_{100.0,100.0}$
 & $92.0_{89.2,94.5}***$
 & $94.5_{92.2,96.5}**$
 \\
gpt-oss\_20b
 & $94.5_{91.0,97.5}**$
 & $\mathbf{100.0}_{100.0,100.0}$
 & $\mathbf{86.0}_{81.0,90.5}$
 & $\mathbf{92.0}_{88.0,95.5}$
 & $\mathbf{8.0}_{4.5,12.0}$
 & $\mathbf{5.5}_{2.5,9.0}$
 & $99.5_{98.8,100.0}$
 & $\mathbf{100.0}_{100.0,100.0}$
 & $97.5_{95.0,99.5}$
 & $99.5_{98.5,100.0}$
 & $\mathbf{99.8}_{99.2,100.0}$
 & $\mathbf{99.0}_{98.0,99.8}$
 \\
nemotron-cascade-2
 & $96.5_{94.0,99.0}$
 & $98.0_{96.0,99.5}$
 & $26.0_{20.0,32.0}***$
 & $42.5_{35.5,49.5}***$
 & $0.0_{0.0,0.0}***$
 & $0.0_{0.0,0.0}**$
 & $\mathbf{100.0}_{100.0,100.0}$
 & $99.8_{99.2,100.0}$
 & $66.5_{60.0,73.0}***$
 & $71.5_{65.0,77.5}***$
 & $83.5_{79.8,87.0}***$
 & $93.0_{90.5,95.5}***$
 \\
qwen3.6\_27b
 & $\mathbf{100.0}_{100.0,100.0}$
 & $\mathbf{100.0}_{100.0,100.0}$
 & $31.5_{25.0,38.0}***$
 & $28.0_{22.0,34.5}***$
 & $0.0_{0.0,0.0}***$
 & $0.0_{0.0,0.0}**$
 & $98.0_{96.5,99.2}$
 & $\mathbf{100.0}_{100.0,100.0}$
 & $95.0_{92.0,98.0}$
 & $98.0_{96.0,99.5}$
 & $88.0_{84.8,91.0}***$
 & $89.2_{86.2,92.2}***$
 \\
\bottomrule
\end{tabular}%
}
\end{table}

%% file: tables/table_main_nonthinking.tex
\begin{table}[t]
\centering
\caption{Main results (non-thinking mode): Exact Match Accuracy (EMA, \%) at 4-shot with 95\% bootstrap CI.
Significance vs.\ best model: $^{*}p{<}0.05$, $^{**}p{<}0.01$, $^{***}p{<}0.001$
(Bonferroni-corrected McNemar's test). ``--'' means that condition was not evaluated. For qwen3.5\_2b on index\_to\_span at 4-shot, the English cell is the observed 113-item subset and the random condition is missing; the combined score is therefore not estimated.}
\label{tab:main_results_nonthinking}
\resizebox{\textwidth}{!}{%
\begin{tabular}{lcccccccccccc}
\toprule
\multicolumn{2}{c}{count\_chars} & \multicolumn{2}{c}{edit\_distance} & \multicolumn{2}{c}{index\_to\_span} & \multicolumn{2}{c}{letter\_container} & \multicolumn{2}{c}{longest\_string\_finder} & \multicolumn{2}{c}{palindrome\_test} \\
\cmidrule(lr){2-3} \cmidrule(lr){4-5} \cmidrule(lr){6-7} \cmidrule(lr){8-9} \cmidrule(lr){10-11} \cmidrule(lr){12-13} & english & random & english & random & english & random & english & random & english & random & english & random \\
\midrule
deepseek\_r1\_32b
 & $76.0_{70.0,82.0}***$
 & $84.5_{79.0,89.5}***$
 & $0.0_{0.0,0.0}***$
 & $0.0_{0.0,0.0}***$
 & $0.0_{0.0,0.0}$
 & $0.0_{0.0,0.0}$
 & $97.8_{96.2,99.0}$
 & $98.2_{97.0,99.5}$
 & $53.0_{46.0,60.0}***$
 & $33.5_{27.0,40.0}***$
 & $87.8_{84.5,90.8}*$
 & $87.8_{84.5,90.8}***$
 \\
gemma4\_31b
 & $94.5_{91.0,97.5}$
 & $98.0_{96.0,99.5}$
 & $\mathbf{39.0}_{32.0,45.5}$
 & $36.0_{29.5,42.5}*$
 & $0.5_{0.0,1.5}$
 & $\mathbf{0.5}_{0.0,1.5}$
 & $\mathbf{98.8}_{97.5,99.8}$
 & $\mathbf{99.8}_{99.2,100.0}$
 & $\mathbf{97.0}_{94.5,99.0}$
 & $\mathbf{100.0}_{100.0,100.0}$
 & $92.2_{89.5,94.8}$
 & $94.2_{91.8,96.5}$
 \\
qwen3.5\_27b
 & $91.5_{87.5,95.0}$
 & $\mathbf{100.0}_{100.0,100.0}$
 & $36.5_{30.0,43.0}$
 & $\mathbf{49.5}_{42.5,56.5}$
 & $\mathbf{1.0}_{0.0,2.5}$
 & $0.0_{0.0,0.0}$
 & $98.5_{97.2,99.5}$
 & $99.2_{98.2,100.0}$
 & $94.0_{90.5,97.0}$
 & $98.5_{96.5,100.0}$
 & $91.2_{88.5,93.8}$
 & $95.2_{93.0,97.2}$
 \\
qwen3.5\_2b
 & $75.5_{69.5,81.5}***$
 & $64.0_{57.5,70.5}***$
 & $23.0_{17.0,29.0}*$
 & $29.0_{23.0,35.5}***$
 & $0.0_{0.0,0.0}$
 & --
 & $77.5_{73.2,81.5}***$
 & $88.0_{84.8,91.2}***$
 & $48.5_{41.5,55.5}***$
 & $47.5_{40.5,54.5}***$
 & $72.0_{67.5,76.2}***$
 & $94.8_{92.5,96.8}$
 \\
qwen3.5\_9b
 & $79.0_{73.0,84.5}***$
 & $97.0_{94.5,99.0}$
 & $34.0_{27.5,40.5}$
 & $45.5_{38.5,52.5}$
 & $0.5_{0.0,1.5}$
 & $\mathbf{0.5}_{0.0,1.5}$
 & $93.5_{91.0,95.8}**$
 & $98.8_{97.5,99.8}$
 & $74.5_{68.0,80.5}***$
 & $89.0_{84.5,93.0}***$
 & $90.5_{87.5,93.2}$
 & $\mathbf{96.8}_{95.0,98.2}$
 \\
qwen3.6\_27b
 & $\mathbf{96.5}_{94.0,99.0}$
 & $99.5_{98.5,100.0}$
 & $32.0_{25.5,38.5}$
 & $33.5_{27.0,40.0}**$
 & $0.0_{0.0,0.0}$
 & $0.0_{0.0,0.0}$
 & $96.8_{95.0,98.5}$
 & $99.5_{98.8,100.0}$
 & $96.5_{93.5,99.0}$
 & $98.5_{96.5,100.0}$
 & $\mathbf{94.2}_{91.8,96.2}$
 & $96.2_{94.2,98.0}$
 \\
\bottomrule
\end{tabular}%
}
\end{table}

%% file: tables/table_shot_ablation.tex
\begin{table}[t]
\centering
\caption{Shot-mode ablation (thinking mode where supported): EMA (\%) across 0-, 1-, and 4-shot settings (condition=all).
\textbf{Bold} = best shot per model $\times$ task.}
\label{tab:shot_ablation}
\resizebox{\textwidth}{!}{%
\begin{tabular}{lcccccccccccccccccc}
\toprule
\multicolumn{3}{c}{count\_chars} & \multicolumn{3}{c}{edit\_distance} & \multicolumn{3}{c}{index\_to\_span} & \multicolumn{3}{c}{letter\_container} & \multicolumn{3}{c}{longest\_string\_finder} & \multicolumn{3}{c}{palindrome\_test} \\
\cmidrule(lr){2-4} \cmidrule(lr){5-7} \cmidrule(lr){8-10} \cmidrule(lr){11-13} \cmidrule(lr){14-16} \cmidrule(lr){17-19} & 0-shot & 1-shot & 4-shot & 0-shot & 1-shot & 4-shot & 0-shot & 1-shot & 4-shot & 0-shot & 1-shot & 4-shot & 0-shot & 1-shot & 4-shot & 0-shot & 1-shot & 4-shot \\
\midrule
deepseek\_r1\_32b
 & 97.8
 & 94.5
 & 96.2
 & 39.8
 & 34.2
 & 37.5
 & 0.5
 & 0.0
 & 0.2
 & 98.9
 & 98.6
 & 99.4
 & 99.8
 & 99.0
 & \textbf{99.8}
 & 95.0
 & 94.2
 & 85.0
 \\
gemma4\_31b
 & 87.8
 & 92.0
 & 96.2
 & 29.0
 & 28.7
 & 39.5
 & 0.5
 & 0.5
 & 0.5
 & 99.5
 & 99.1
 & 99.2
 & 95.8
 & 97.8
 & 98.5
 & 84.1
 & 92.9
 & 93.2
 \\
gpt-oss\_20b
 & 96.5
 & 97.5
 & 97.2
 & \textbf{85.2}
 & \textbf{88.8}
 & \textbf{89.0}
 & \textbf{7.4}
 & \textbf{6.5}
 & \textbf{6.8}
 & 99.5
 & 99.2
 & 99.8
 & 99.8
 & \textbf{99.8}
 & 98.5
 & \textbf{98.8}
 & \textbf{98.9}
 & \textbf{99.4}
 \\
nemotron-cascade-2
 & 96.2
 & 96.2
 & 97.2
 & 23.0
 & 16.0
 & 34.2
 & 0.0
 & 0.0
 & 0.0
 & 99.8
 & 99.5
 & \textbf{99.9}
 & 57.5
 & 72.8
 & 69.0
 & 81.9
 & 86.8
 & 88.2
 \\
qwen3.6\_27b
 & \textbf{99.2}
 & \textbf{99.8}
 & \textbf{100.0}
 & 37.2
 & 35.8
 & 29.8
 & 0.0
 & 0.0
 & 0.0
 & \textbf{99.9}
 & \textbf{99.6}
 & 99.0
 & \textbf{100.0}
 & 99.3
 & 96.5
 & 95.5
 & 81.9
 & 88.6
 \\
\bottomrule
\end{tabular}%
}
\end{table}

%% file: tables/table_kappa.tex
\begin{table}[t]
\centering
\caption{Cohen's $\kappa$ (thinking mode where supported) (4-shot, condition=all). Annotation uses headline $\kappa$ (linear weighted for integer tasks).}
\label{tab:kappa}
\resizebox{\textwidth}{!}{%
\begin{tabular}{lccccc}
\hline
Model  & count\_chars & edit\_distance & letter\_container & longest\_string\_finder & palindrome\_test \\
\hline
deepseek\_r1\_32b
 & $0.984$ {\scriptsize almost\_perfect}
 & $0.556$ {\scriptsize moderate}
 & $0.988$ {\scriptsize almost\_perfect}
 & $0.997$ {\scriptsize almost\_perfect}
 & $0.7$ {\scriptsize substantial}
 \\
gemma4\_31b
 & $0.988$ {\scriptsize almost\_perfect}
 & $0.743$ {\scriptsize substantial}
 & $0.985$ {\scriptsize almost\_perfect}
 & $0.985$ {\scriptsize almost\_perfect}
 & $0.865$ {\scriptsize almost\_perfect}
 \\
gpt-oss\_20b
 & $0.991$ {\scriptsize almost\_perfect}
 & $0.941$ {\scriptsize almost\_perfect}
 & $0.995$ {\scriptsize almost\_perfect}
 & $0.985$ {\scriptsize almost\_perfect}
 & $0.988$ {\scriptsize almost\_perfect}
 \\
nemotron-cascade-2
 & $0.988$ {\scriptsize almost\_perfect}
 & $0.623$ {\scriptsize substantial}
 & $0.998$ {\scriptsize almost\_perfect}
 & $0.689$ {\scriptsize substantial}
 & $0.765$ {\scriptsize substantial}
 \\
qwen3.6\_27b
 & $1.0$ {\scriptsize almost\_perfect}
 & $0.563$ {\scriptsize moderate}
 & $0.98$ {\scriptsize almost\_perfect}
 & $0.965$ {\scriptsize almost\_perfect}
 & $0.772$ {\scriptsize substantial}
 \\
\hline
\end{tabular}%
}
\end{table}

%% file: tables/table_significance_matrix.tex
\begin{table}[t]
\centering
\caption{McNemar pairwise comparisons (thinking mode where supported): fraction of Bonferroni-significant task×shot cells (numerator/denominator). Blue tint: row model tends to win net; red: column model; gray: mixed.}
\label{tab:mcnemar_sig_matrix}
\resizebox{\textwidth}{!}{%
\begin{tabular}{lccccc}
\hline
 &  deepseek\_r1\_32b &  gemma4\_31b &  gpt-oss\_20b &  nemotron-cascade-2 &  qwen3.6\_27b \\
\hline
deepseek\_r1\_32b
 & ---
 & \cellcolor{blue!25}{$5/18$}
 & \cellcolor{red!25}{$9/18$}
 & \cellcolor{blue!25}{$7/18$}
 & \cellcolor{gray!15}{$4/18$}
 \\
gemma4\_31b
 & \cellcolor{red!25}{$5/18$}
 & ---
 & \cellcolor{red!25}{$12/18$}
 & \cellcolor{blue!25}{$7/18$}
 & \cellcolor{red!25}{$8/18$}
 \\
gpt-oss\_20b
 & \cellcolor{blue!25}{$9/18$}
 & \cellcolor{blue!25}{$12/18$}
 & ---
 & \cellcolor{blue!25}{$12/18$}
 & \cellcolor{blue!25}{$10/18$}
 \\
nemotron-cascade-2
 & \cellcolor{red!25}{$7/18$}
 & \cellcolor{red!25}{$7/18$}
 & \cellcolor{red!25}{$12/18$}
 & ---
 & \cellcolor{red!25}{$8/18$}
 \\
qwen3.6\_27b
 & \cellcolor{gray!15}{$4/18$}
 & \cellcolor{blue!25}{$8/18$}
 & \cellcolor{red!25}{$10/18$}
 & \cellcolor{blue!25}{$8/18$}
 & ---
 \\
\hline
\end{tabular}%
}
\end{table}

%% file: tables/table_significance_matrix_nonthinking.tex
\begin{table}[t]
\centering
\caption{McNemar pairwise comparisons (non-thinking mode): fraction of Bonferroni-significant task×shot cells (numerator/denominator). Blue tint: row model tends to win net; red: column model; gray: mixed.}
\label{tab:mcnemar_sig_matrix_nonthinking}
\resizebox{\textwidth}{!}{%
\begin{tabular}{lcccccc}
\hline
 &  deepseek\_r1\_32b &  gemma4\_31b &  qwen3.5\_27b &  qwen3.5\_2b &  qwen3.5\_9b &  qwen3.6\_27b \\
\hline
deepseek\_r1\_32b
 & ---
 & \cellcolor{red!25}{$9/18$}
 & \cellcolor{red!25}{$9/18$}
 & \cellcolor{blue!25}{$11/18$}
 & \cellcolor{gray!15}{$12/18$}
 & \cellcolor{red!25}{$9/18$}
 \\
gemma4\_31b
 & \cellcolor{blue!25}{$9/18$}
 & ---
 & \cellcolor{red!25}{$6/18$}
 & \cellcolor{blue!25}{$15/18$}
 & \cellcolor{blue!25}{$8/18$}
 & \cellcolor{red!25}{$5/18$}
 \\
qwen3.5\_27b
 & \cellcolor{blue!25}{$9/18$}
 & \cellcolor{blue!25}{$6/18$}
 & ---
 & \cellcolor{blue!25}{$15/18$}
 & \cellcolor{blue!25}{$12/18$}
 & \cellcolor{blue!25}{$3/18$}
 \\
qwen3.5\_2b
 & \cellcolor{red!25}{$11/18$}
 & \cellcolor{red!25}{$15/18$}
 & \cellcolor{red!25}{$15/18$}
 & ---
 & \cellcolor{red!25}{$14/18$}
 & \cellcolor{red!25}{$14/18$}
 \\
qwen3.5\_9b
 & \cellcolor{gray!15}{$12/18$}
 & \cellcolor{red!25}{$8/18$}
 & \cellcolor{red!25}{$12/18$}
 & \cellcolor{blue!25}{$14/18$}
 & ---
 & \cellcolor{red!25}{$9/18$}
 \\
qwen3.6\_27b
 & \cellcolor{blue!25}{$9/18$}
 & \cellcolor{blue!25}{$5/18$}
 & \cellcolor{red!25}{$3/18$}
 & \cellcolor{blue!25}{$14/18$}
 & \cellcolor{blue!25}{$9/18$}
 & ---
 \\
\hline
\end{tabular}%
}
\end{table}

%% file: tables/table_kendall_tau.tex
\begin{table}[t]
\centering
\caption{Kendall's $\tau$ (thinking mode where supported) between shot count $\{0,1,4\}$ and EMA (condition=all). Low power at $n{=}3$ points — interpret patterns qualitatively.}
\label{tab:kendall_tau}
\resizebox{\textwidth}{!}{%
\begin{tabular}{lcccccc}
\hline
Model  & count\_chars & edit\_distance & index\_to\_span & letter\_container & longest\_string\_finder & palindrome\_test \\
\hline
deepseek\_r1\_32b
 & $\tau{=}-0.333$ {\scriptsize valley\_1shot}
 & $\tau{=}-0.333$ {\scriptsize valley\_1shot}
 & $\tau{=}-0.333$ {\scriptsize valley\_1shot}
 & $\tau{=}0.333$ {\scriptsize valley\_1shot}
 & $\tau{=}0.0$ {\scriptsize valley\_1shot}
 & $\tau{=}-1.0$ {\scriptsize monotone\_down}
 \\
gemma4\_31b
 & $\tau{=}1.0$ {\scriptsize monotone\_up}
 & $\tau{=}0.333$ {\scriptsize valley\_1shot}
 & --
 & $\tau{=}-0.333$ {\scriptsize valley\_1shot}
 & $\tau{=}1.0$ {\scriptsize monotone\_up}
 & $\tau{=}1.0$ {\scriptsize monotone\_up}
 \\
gpt-oss\_20b
 & $\tau{=}0.333$ {\scriptsize peak\_1shot}
 & $\tau{=}1.0$ {\scriptsize monotone\_up}
 & $\tau{=}-0.333$ {\scriptsize valley\_1shot}
 & $\tau{=}0.333$ {\scriptsize valley\_1shot}
 & $\tau{=}-0.816$ {\scriptsize monotone\_down}
 & $\tau{=}1.0$ {\scriptsize monotone\_up}
 \\
nemotron-cascade-2
 & $\tau{=}0.816$ {\scriptsize monotone\_up}
 & $\tau{=}0.333$ {\scriptsize valley\_1shot}
 & --
 & $\tau{=}0.333$ {\scriptsize valley\_1shot}
 & $\tau{=}0.333$ {\scriptsize peak\_1shot}
 & $\tau{=}1.0$ {\scriptsize monotone\_up}
 \\
qwen3.6\_27b
 & $\tau{=}1.0$ {\scriptsize monotone\_up}
 & $\tau{=}-1.0$ {\scriptsize monotone\_down}
 & --
 & $\tau{=}-1.0$ {\scriptsize monotone\_down}
 & $\tau{=}-1.0$ {\scriptsize monotone\_down}
 & $\tau{=}-0.333$ {\scriptsize valley\_1shot}
 \\
\hline
\end{tabular}%
}
\end{table}

%% file: tables/table_eng_vs_rand.tex
\begin{table}[t]
\centering
\caption{English vs.\ random gap (thinking mode where supported) ($\Delta$EMA = EMA$_\text{eng}$ $-$ EMA$_\text{rand}$, \%) at 4-shot.
\colorbox{posblue}{\strut Blue} = English advantage; \colorbox{negred}{\strut Red} = random advantage.
Significance (Bonferroni-corrected): $^{*}p{<}0.05$, $^{**}p{<}0.01$, $^{***}p{<}0.001$.}
\label{tab:eng_vs_rand}
\resizebox{\textwidth}{!}{%
\begin{tabular}{lcccccc}
\toprule
& count\_chars & edit\_distance & index\_to\_span & letter\_container & longest\_string\_finder & palindrome\_test \\
\midrule
deepseek\_r1\_32b
 & \cellcolor{negred}$-7.5*$
 & \cellcolor{negred}$-6.0$
 & \cellcolor{negred}$-0.5$
 & \cellcolor{negred}$-1.2$
 & \cellcolor{negred}$-0.5$
 & \cellcolor{negred}$-3.5$
 \\
gemma4\_31b
 & \cellcolor{negred}$-3.5$
 & \cellcolor{posblue}$+10.0$
 & $0.0$
 & \cellcolor{negred}$-1.0$
 & \cellcolor{negred}$-3.0$
 & \cellcolor{negred}$-2.5$
 \\
gpt-oss\_20b
 & \cellcolor{negred}$-5.5$
 & \cellcolor{negred}$-6.0$
 & \cellcolor{posblue}$+2.5$
 & \cellcolor{negred}$-0.5$
 & \cellcolor{negred}$-2.0$
 & \cellcolor{posblue}$+0.8$
 \\
nemotron-cascade-2
 & \cellcolor{negred}$-1.5$
 & \cellcolor{negred}$-16.5$
 & $0.0$
 & \cellcolor{posblue}$+0.2$
 & \cellcolor{negred}$-5.0$
 & \cellcolor{negred}$-9.5**$
 \\
qwen3.6\_27b
 & $0.0$
 & \cellcolor{posblue}$+3.5$
 & $0.0$
 & \cellcolor{negred}$-2.0$
 & \cellcolor{negred}$-3.0$
 & \cellcolor{negred}$-1.2$
 \\
\bottomrule
\end{tabular}%
}
\end{table}

%% file: tables/table_parse_errors.tex
\begin{table}[t]
\centering
\caption{Parse error rate (PER, \%) across models, tasks, and shot settings (thinking mode where supported) (condition=all).
Format: PER (\textit{blank}; \textit{inf.err}).}
\label{tab:parse_errors}
\resizebox{\textwidth}{!}{%
\begin{tabular}{lcccccccccccccccccc}
\toprule
\multicolumn{3}{c}{count\_chars} & \multicolumn{3}{c}{edit\_distance} & \multicolumn{3}{c}{index\_to\_span} & \multicolumn{3}{c}{letter\_container} & \multicolumn{3}{c}{longest\_string\_finder} & \multicolumn{3}{c}{palindrome\_test} \\
\cmidrule(lr){2-4} \cmidrule(lr){5-7} \cmidrule(lr){8-10} \cmidrule(lr){11-13} \cmidrule(lr){14-16} \cmidrule(lr){17-19} & 0-shot & 1-shot & 4-shot & 0-shot & 1-shot & 4-shot & 0-shot & 1-shot & 4-shot & 0-shot & 1-shot & 4-shot & 0-shot & 1-shot & 4-shot & 0-shot & 1-shot & 4-shot \\
\midrule
deepseek\_r1\_32b
 & 0.0\,(\textit{0.0};\,\textit{0.0})
 & 0.0\,(\textit{0.0};\,\textit{0.0})
 & 0.0\,(\textit{0.0};\,\textit{0.0})
 & 0.5\,(\textit{0.0};\,\textit{0.5})
 & 0.2\,(\textit{0.0};\,\textit{0.2})
 & 0.5\,(\textit{0.0};\,\textit{0.5})
 & 0.2\,(\textit{0.0};\,\textit{0.2})
 & 0.2\,(\textit{0.0};\,\textit{0.2})
 & 0.5\,(\textit{0.0};\,\textit{0.5})
 & 0.0\,(\textit{0.0};\,\textit{0.0})
 & 0.0\,(\textit{0.0};\,\textit{0.0})
 & 0.0\,(\textit{0.0};\,\textit{0.0})
 & 0.2\,(\textit{0.0};\,\textit{0.2})
 & 0.0\,(\textit{0.0};\,\textit{0.0})
 & 0.2\,(\textit{0.0};\,\textit{0.2})
 & 3.5\,(\textit{0.0};\,\textit{3.5})
 & 4.5\,(\textit{0.0};\,\textit{4.5})
 & 14.8\,(\textit{0.0};\,\textit{14.8})
 \\
gemma4\_31b
 & 0.0\,(\textit{0.0};\,\textit{0.0})
 & 0.0\,(\textit{0.0};\,\textit{0.0})
 & 0.0\,(\textit{0.0};\,\textit{0.0})
 & 0.0\,(\textit{0.0};\,\textit{0.0})
 & 0.0\,(\textit{0.0};\,\textit{0.0})
 & 0.0\,(\textit{0.0};\,\textit{0.0})
 & 0.0\,(\textit{0.0};\,\textit{0.0})
 & 0.0\,(\textit{0.0};\,\textit{0.0})
 & 0.0\,(\textit{0.0};\,\textit{0.0})
 & 0.0\,(\textit{0.0};\,\textit{0.0})
 & 0.0\,(\textit{0.0};\,\textit{0.0})
 & 0.0\,(\textit{0.0};\,\textit{0.0})
 & 0.0\,(\textit{0.0};\,\textit{0.0})
 & 0.0\,(\textit{0.0};\,\textit{0.0})
 & 0.0\,(\textit{0.0};\,\textit{0.0})
 & 0.0\,(\textit{0.0};\,\textit{0.0})
 & 0.0\,(\textit{0.0};\,\textit{0.0})
 & 0.0\,(\textit{0.0};\,\textit{0.0})
 \\
gpt-oss\_20b
 & 0.0\,(\textit{0.0};\,\textit{0.0})
 & 0.0\,(\textit{0.0};\,\textit{0.0})
 & 0.0\,(\textit{0.0};\,\textit{0.0})
 & 2.1\,(\textit{0.0};\,\textit{0.0})
 & 0.0\,(\textit{0.0};\,\textit{0.0})
 & 0.0\,(\textit{0.0};\,\textit{0.0})
 & 3.7\,(\textit{0.0};\,\textit{3.7})
 & 17.2\,(\textit{0.0};\,\textit{17.2})
 & 21.2\,(\textit{0.0};\,\textit{21.2})
 & 0.0\,(\textit{0.0};\,\textit{0.0})
 & 0.0\,(\textit{0.0};\,\textit{0.0})
 & 0.0\,(\textit{0.0};\,\textit{0.0})
 & 0.0\,(\textit{0.0};\,\textit{0.0})
 & 0.2\,(\textit{0.0};\,\textit{0.0})
 & 0.5\,(\textit{0.0};\,\textit{0.0})
 & 0.0\,(\textit{0.0};\,\textit{0.0})
 & 0.0\,(\textit{0.0};\,\textit{0.0})
 & 0.0\,(\textit{0.0};\,\textit{0.0})
 \\
nemotron-cascade-2
 & 0.0\,(\textit{0.0};\,\textit{0.0})
 & 0.0\,(\textit{0.0};\,\textit{0.0})
 & 0.0\,(\textit{0.0};\,\textit{0.0})
 & 2.0\,(\textit{0.0};\,\textit{0.0})
 & 0.0\,(\textit{0.0};\,\textit{0.0})
 & 0.8\,(\textit{0.0};\,\textit{0.0})
 & 0.3\,(\textit{0.0};\,\textit{0.3})
 & 2.2\,(\textit{0.0};\,\textit{2.2})
 & 2.2\,(\textit{0.0};\,\textit{2.2})
 & 0.0\,(\textit{0.0};\,\textit{0.0})
 & 0.0\,(\textit{0.0};\,\textit{0.0})
 & 0.0\,(\textit{0.0};\,\textit{0.0})
 & 15.2\,(\textit{0.0};\,\textit{0.0})
 & 0.0\,(\textit{0.0};\,\textit{0.0})
 & 1.2\,(\textit{0.0};\,\textit{0.0})
 & 4.4\,(\textit{0.0};\,\textit{0.0})
 & 0.0\,(\textit{0.0};\,\textit{0.0})
 & 0.4\,(\textit{0.0};\,\textit{0.0})
 \\
qwen3.6\_27b
 & 0.0\,(\textit{0.0};\,\textit{0.0})
 & 0.0\,(\textit{0.0};\,\textit{0.0})
 & 0.0\,(\textit{0.0};\,\textit{0.0})
 & 0.0\,(\textit{0.0};\,\textit{0.0})
 & 0.0\,(\textit{0.0};\,\textit{0.0})
 & 0.0\,(\textit{0.0};\,\textit{0.0})
 & 0.0\,(\textit{0.0};\,\textit{0.0})
 & 0.0\,(\textit{0.0};\,\textit{0.0})
 & 0.0\,(\textit{0.0};\,\textit{0.0})
 & 0.0\,(\textit{0.0};\,\textit{0.0})
 & 0.0\,(\textit{0.0};\,\textit{0.0})
 & 0.0\,(\textit{0.0};\,\textit{0.0})
 & 0.0\,(\textit{0.0};\,\textit{0.0})
 & 0.0\,(\textit{0.0};\,\textit{0.0})
 & 0.0\,(\textit{0.0};\,\textit{0.0})
 & 0.0\,(\textit{0.0};\,\textit{0.0})
 & 0.0\,(\textit{0.0};\,\textit{0.0})
 & 0.0\,(\textit{0.0};\,\textit{0.0})
 \\
\bottomrule
\end{tabular}%
}
\end{table}

%% file: tables/table_delta_dist.tex
\begin{table}[t]
\centering
\caption{Signed delta distribution ($\hat{y} - y$) for \texttt{count\_chars} (thinking mode where supported) at 4-shot (valid responses only).}
\label{tab:delta_dist_count_chars}
\begin{tabular}{llrrrrr}
\toprule
Model & Cond. & Mean $\pm$ Std & Median & P5 & P95 & $n$ \\
\midrule
\multirow{3}{*}{deepseek\_r1\_32b}
 & english
 & $0.03 \pm 0.39$ & 0.00 & 0.00 & 0.05 & 200 \\
 & random
 & $0.00 \pm 0.00$ & 0.00 & 0.00 & 0.00 & 200 \\
 & all
 & $0.01 \pm 0.27$ & 0.00 & 0.00 & 0.00 & 400 \\
\midrule\multirow{3}{*}{gemma4\_31b}
 & english
 & $0.06 \pm 0.23$ & 0.00 & 0.00 & 1.00 & 200 \\
 & random
 & $0.01 \pm 0.14$ & 0.00 & 0.00 & 0.00 & 200 \\
 & all
 & $0.03 \pm 0.19$ & 0.00 & 0.00 & 0.00 & 400 \\
\midrule\multirow{3}{*}{gpt-oss\_20b}
 & english
 & $-0.05 \pm 0.26$ & 0.00 & -0.05 & 0.00 & 200 \\
 & random
 & $0.00 \pm 0.00$ & 0.00 & 0.00 & 0.00 & 200 \\
 & all
 & $-0.03 \pm 0.19$ & 0.00 & 0.00 & 0.00 & 400 \\
\midrule\multirow{3}{*}{nemotron-cascade-2}
 & english
 & $-0.03 \pm 0.31$ & 0.00 & 0.00 & 0.00 & 200 \\
 & random
 & $0.00 \pm 0.14$ & 0.00 & 0.00 & 0.00 & 200 \\
 & all
 & $-0.01 \pm 0.24$ & 0.00 & 0.00 & 0.00 & 400 \\
\midrule\multirow{3}{*}{qwen3.6\_27b}
 & english
 & $0.00 \pm 0.00$ & 0.00 & 0.00 & 0.00 & 200 \\
 & random
 & $0.00 \pm 0.00$ & 0.00 & 0.00 & 0.00 & 200 \\
 & all
 & $0.00 \pm 0.00$ & 0.00 & 0.00 & 0.00 & 400 \\
\bottomrule
\end{tabular}
\end{table}

\begin{table}[t]
\centering
\caption{Signed delta distribution ($\hat{y} - y$) for \texttt{edit\_distance} (thinking mode where supported) at 4-shot (valid responses only).}
\label{tab:delta_dist_edit_distance}
\begin{tabular}{llrrrrr}
\toprule
Model & Cond. & Mean $\pm$ Std & Median & P5 & P95 & $n$ \\
\midrule
\multirow{3}{*}{deepseek\_r1\_32b}
 & english
 & $0.00 \pm 1.59$ & 0.00 & -2.00 & 3.00 & 199 \\
 & random
 & $-0.07 \pm 2.15$ & 0.00 & -3.00 & 3.10 & 199 \\
 & all
 & $-0.04 \pm 1.89$ & 0.00 & -2.15 & 3.00 & 398 \\
\midrule\multirow{3}{*}{gemma4\_31b}
 & english
 & $0.01 \pm 0.92$ & 0.00 & -1.00 & 1.05 & 200 \\
 & random
 & $-0.66 \pm 0.73$ & -1.00 & -2.00 & 0.00 & 200 \\
 & all
 & $-0.33 \pm 0.90$ & 0.00 & -2.00 & 1.00 & 400 \\
\midrule\multirow{3}{*}{gpt-oss\_20b}
 & english
 & $0.07 \pm 0.74$ & 0.00 & -1.00 & 1.00 & 200 \\
 & random
 & $0.00 \pm 0.40$ & 0.00 & 0.00 & 0.00 & 200 \\
 & all
 & $0.03 \pm 0.60$ & 0.00 & -1.00 & 1.00 & 400 \\
\midrule\multirow{3}{*}{nemotron-cascade-2}
 & english
 & $-0.06 \pm 1.33$ & 0.00 & -2.00 & 2.00 & 197 \\
 & random
 & $-0.40 \pm 1.06$ & 0.00 & -2.00 & 1.00 & 200 \\
 & all
 & $-0.23 \pm 1.21$ & 0.00 & -2.00 & 2.00 & 397 \\
\midrule\multirow{3}{*}{qwen3.6\_27b}
 & english
 & $-0.69 \pm 1.29$ & -1.00 & -3.00 & 1.00 & 200 \\
 & random
 & $-1.01 \pm 1.14$ & -1.00 & -3.00 & 1.00 & 200 \\
 & all
 & $-0.85 \pm 1.23$ & -1.00 & -3.00 & 1.00 & 400 \\
\bottomrule
\end{tabular}
\end{table}

%% file: tables/table_thinking_cost.tex
\begin{table}[t]
\centering
\caption{Thinking-cost efficiency at 4-shot (thinking mode, condition=all). Top-3 per task by EMA / mean thinking tokens (higher is better). NaN tokens (no paired runs) shown as ``--''.}
\label{tab:thinking-cost-efficiency}
\begin{tabular}{llrrr}
\hline
Task & Model & EMA & Mean tokens & EMA/token \\
\hline
count\_chars
 & deepseek\_r1\_32b
 & 96.2\%
 & 274 & $0.00351$ \\
count\_chars
 & gemma4\_31b
 & 96.2\%
 & 0 & -- \\
count\_chars
 & gpt-oss\_20b
 & 97.2\%
 & -- & -- \\
edit\_distance
 & deepseek\_r1\_32b
 & 37.5\%
 & 1555 & $0.00024$ \\
edit\_distance
 & gemma4\_31b
 & 39.5\%
 & 0 & -- \\
edit\_distance
 & gpt-oss\_20b
 & 89.0\%
 & 0 & -- \\
index\_to\_span
 & deepseek\_r1\_32b
 & 0.2\%
 & 381 & $1e-05$ \\
index\_to\_span
 & gemma4\_31b
 & 0.5\%
 & 0 & -- \\
index\_to\_span
 & gpt-oss\_20b
 & 6.8\%
 & 0 & -- \\
letter\_container
 & deepseek\_r1\_32b
 & 99.4\%
 & 202 & $0.00493$ \\
letter\_container
 & gemma4\_31b
 & 99.2\%
 & 0 & -- \\
letter\_container
 & gpt-oss\_20b
 & 99.8\%
 & -- & -- \\
longest\_string\_finder
 & deepseek\_r1\_32b
 & 99.8\%
 & 328 & $0.00304$ \\
longest\_string\_finder
 & gemma4\_31b
 & 98.5\%
 & 0 & -- \\
longest\_string\_finder
 & gpt-oss\_20b
 & 98.5\%
 & 0 & -- \\
palindrome\_test
 & deepseek\_r1\_32b
 & 85.0\%
 & 228 & $0.00373$ \\
palindrome\_test
 & gemma4\_31b
 & 93.2\%
 & 0 & -- \\
palindrome\_test
 & gpt-oss\_20b
 & 99.4\%
 & 0 & -- \\
\hline
\end{tabular}
\end{table}

%% file: tables/table_effect_sizes.tex
\begin{table}[t]
\centering
\caption{Top-12 model-pair effect sizes at 4-shot (thinking, condition=all), sorted by $|h|$. CIs are 95\% bootstrap percentile.}
\label{tab:effect-sizes}
\resizebox{\textwidth}{!}{%
\begin{tabular}{llllrrr}
\hline
Task & Model A & Model B & $h$ [95\% CI] & Risk diff. & $p$ & $q$ \\
\hline
edit\_distance
 & gpt-oss\_20b
 & qwen3.6\_27b
 & $1.31$ [1.18, 1.45]
 & $0.593$ & $0.0$ & $0.0$ \\
edit\_distance
 & gpt-oss\_20b
 & nemotron-cascade-2
 & $1.22$ [1.09, 1.36]
 & $0.547$ & $0.0$ & $0.0$ \\
edit\_distance
 & deepseek\_r1\_32b
 & gpt-oss\_20b
 & $-1.15$ [-1.29, -1.01]
 & $-0.515$ & $0.0$ & $0.0$ \\
edit\_distance
 & gemma4\_31b
 & gpt-oss\_20b
 & $-1.11$ [-1.25, -0.97]
 & $-0.495$ & $0.0$ & $0.0$ \\
longest\_string\_finder
 & deepseek\_r1\_32b
 & nemotron-cascade-2
 & $1.08$ [0.95, 1.26]
 & $0.308$ & $0.0$ & $0.0$ \\
longest\_string\_finder
 & gemma4\_31b
 & nemotron-cascade-2
 & $0.94$ [0.8, 1.09]
 & $0.295$ & $0.0$ & $0.0$ \\
longest\_string\_finder
 & gpt-oss\_20b
 & nemotron-cascade-2
 & $0.94$ [0.8, 1.09]
 & $0.295$ & $0.0$ & $0.0$ \\
longest\_string\_finder
 & nemotron-cascade-2
 & qwen3.6\_27b
 & $-0.8$ [-0.94, -0.67]
 & $-0.275$ & $0.0$ & $0.0$ \\
palindrome\_test
 & deepseek\_r1\_32b
 & gpt-oss\_20b
 & $-0.64$ [-0.74, -0.55]
 & $-0.144$ & $0.0$ & $0.0$ \\
palindrome\_test
 & gpt-oss\_20b
 & nemotron-cascade-2
 & $0.54$ [0.45, 0.65]
 & $0.111$ & $0.0$ & $0.0$ \\
palindrome\_test
 & gpt-oss\_20b
 & qwen3.6\_27b
 & $0.53$ [0.43, 0.64]
 & $0.108$ & $0.0$ & $0.0$ \\
index\_to\_span
 & gpt-oss\_20b
 & nemotron-cascade-2
 & $0.53$ [0.42, 0.62]
 & $0.068$ & $0.0$ & $0.0$ \\
\hline
\end{tabular}%
}
\end{table}

%% file: tables/table_composite_leaderboard.tex
\begin{table}[t]
\centering
\caption{Composite leaderboard at 4-shot (thinking, condition=all). Macro EMA averages over complete task-level EMAs and omits a task when either input condition is missing; CI is 95\% bootstrap percentile of those task-level EMAs.}
\label{tab:composite_leaderboard_full}
\begin{tabular}{rlrrr}
\hline
Rank & Model & Macro EMA [95\% CI] & VRR & PER \\
\hline
1
 & gpt-oss\_20b
 & $0.818$ [0.512, 0.989]
 & $0.964$ & $0.036$ \\
2
 & gemma4\_31b
 & $0.712$ [0.388, 0.971]
 & $1.0$ & $0.0$ \\
3
 & deepseek\_r1\_32b
 & $0.697$ [0.385, 0.966]
 & $0.973$ & $0.027$ \\
4
 & qwen3.6\_27b
 & $0.69$ [0.358, 0.972]
 & $1.0$ & $0.0$ \\
5
 & nemotron-cascade-2
 & $0.648$ [0.339, 0.919]
 & $0.992$ & $0.008$ \\
\hline
\end{tabular}
\end{table}

%% file: tables/table_composite_leaderboard_nonthinking.tex
\begin{table}[t]
\centering
\caption{Composite leaderboard at 4-shot (non-thinking mode). Macro EMA averages over complete task-level EMAs and omits a task when either input condition is missing; CI is 95\% bootstrap percentile of those task-level EMAs.}
\label{tab:composite_leaderboard_nonthinking}
\begin{tabular}{rlrrr}
\hline
Rank & Model & Macro EMA [95\% CI] & VRR & PER \\
\hline
1
 & qwen3.5\_27b
 & $0.713$ [0.394, 0.966]
 & $1.0$ & $0.0$ \\
2
 & gemma4\_31b
 & $0.709$ [0.385, 0.971]
 & $1.0$ & $0.0$ \\
3
 & qwen3.6\_27b
 & $0.703$ [0.376, 0.975]
 & $1.0$ & $0.0$ \\
4
 & qwen3.5\_9b
 & $0.666$ [0.361, 0.915]
 & $0.999$ & $0.001$ \\
5
 & qwen3.5\_2b
 & $0.62$ [0.419, 0.804]
 & $1.0$ & $0.0$ \\
6
 & deepseek\_r1\_32b
 & $0.515$ [0.163, 0.83]
 & $0.623$ & $0.377$ \\
\hline
\end{tabular}
\end{table}